\documentclass[english]{article}
\usepackage[T1]{fontenc}
\usepackage[latin9]{inputenc}
\usepackage{geometry}
\usepackage{color}
\usepackage{xcolor}
\usepackage{url}
\usepackage{babel}
\usepackage{booktabs}
\usepackage{multirow}
\usepackage{makecell}
\usepackage{siunitx}
\usepackage{mathtools}
\usepackage{bm}
\usepackage{bbm}
\usepackage{amsmath}
\usepackage{amsthm}
\usepackage{amssymb}
\usepackage{algorithm}
\usepackage{algorithmic}
\usepackage{graphicx}
\usepackage{caption}
\usepackage{subcaption}
\usepackage[unicode=true,pdfusetitle,
 bookmarks=true,bookmarksnumbered=true,bookmarksopen=true,bookmarksopenlevel=1,
 breaklinks=false,pdfborder={0 0 0},pdfborderstyle={},backref=false,colorlinks=true]
 {hyperref}
\hypersetup{
 linkcolor=blue, urlcolor=blue, citecolor=blue, linktoc=all, linktocpage=false}
\hypersetup{bookmarksopen=true,
        bookmarksopenlevel=1, 
        bookmarksnumbered=true, }
\usepackage{appendix}
\usepackage{xspace}
\usepackage{verbatim}
\usepackage{subcaption}

\usepackage{listings}
\definecolor{codegreen}{rgb}{0,0.6,0}
\definecolor{codegray}{rgb}{0.5,0.5,0.5}
\definecolor{codepurple}{rgb}{0.58,0,0.82}
\definecolor{backcolour}{rgb}{0.98,0.98,0.98}
\lstdefinestyle{mystyle}{
    backgroundcolor=\color{backcolour},   
    commentstyle=\color{codegreen},
    keywordstyle=\color{magenta},
    numberstyle=\tiny\color{codegray},
    stringstyle=\color{codepurple},
    basicstyle=\ttfamily\small,
    breakatwhitespace=false,         
    breaklines=true,                 
    captionpos=b,                    
    keepspaces=true,                 
    numbers=left,                    
    numbersep=5pt,                  
    showspaces=false,                
    showstringspaces=false,
    showtabs=false,                  
    tabsize=2
}
\makeatletter

\theoremstyle{plain}

\theoremstyle{remark}

\theoremstyle{plain}

\theoremstyle{plain}

\theoremstyle{plain}

\theoremstyle{plain}

\theoremstyle{plain}

\theoremstyle{definition}

\theoremstyle{definition}

\theoremstyle{plain}

\definecolor{todo}{RGB}{200,0,20}

\definecolor{new}{RGB}{200,0,200}

\definecolor{emerald}{rgb}{0.31, 0.78, 0.47}

\allowdisplaybreaks

\newcommand{\trinity}{\texttt{Trinity-RFT}\xspace}
\newcommand{\trinitystudio}{\texttt{Trinity-Studio}\xspace}

\newcommand{\mix}{\texttt{MIX}\xspace}

\definecolor{codegreen}{rgb}{0,0.6,0}
\definecolor{codegray}{rgb}{0.5,0.5,0.5}
\definecolor{codepurple}{rgb}{0.58,0,0.82}
\definecolor{backcolour}{rgb}{0.95,0.95,0.92}

\lstdefinelanguage{YAML}{
  keywords={true,false,null,yes,no},
  keywordstyle=\color{darkgray}\bfseries,
  basicstyle=\ttfamily\small,
  ndkeywords={apiVersion,data,kind,metadata,name,namespace,spec,replicas,selector,template,containers,image,ports,containerPort},
  ndkeywordstyle=\color{codepurple}\bfseries,
  comment=[l]{\#},
  commentstyle=\color{codegreen},
  stringstyle=\color{red},
  string=[b]',
  string=[b]",
  morestring=[b]`,
  moredelim=**[s][\color{blue}]{:}{\ }, %
  sensitive=true,
  morecomment=[l][\color{magenta}]{---}, %
}

\makeatother

\providecommand{\assumptionname}{Assumption}
\providecommand{\corollaryname}{Corollary}
\providecommand{\examplename}{Example}
\providecommand{\definitionname}{Definition}
\providecommand{\factname}{Fact}
\providecommand{\lemmaname}{Lemma}
\providecommand{\propositionname}{Proposition}
\providecommand{\remarkname}{Remark}
\providecommand{\theoremname}{Theorem}

\begin{document}
\title{
  \raisebox{-.4\height}{\includegraphics[width=0.18\textwidth]{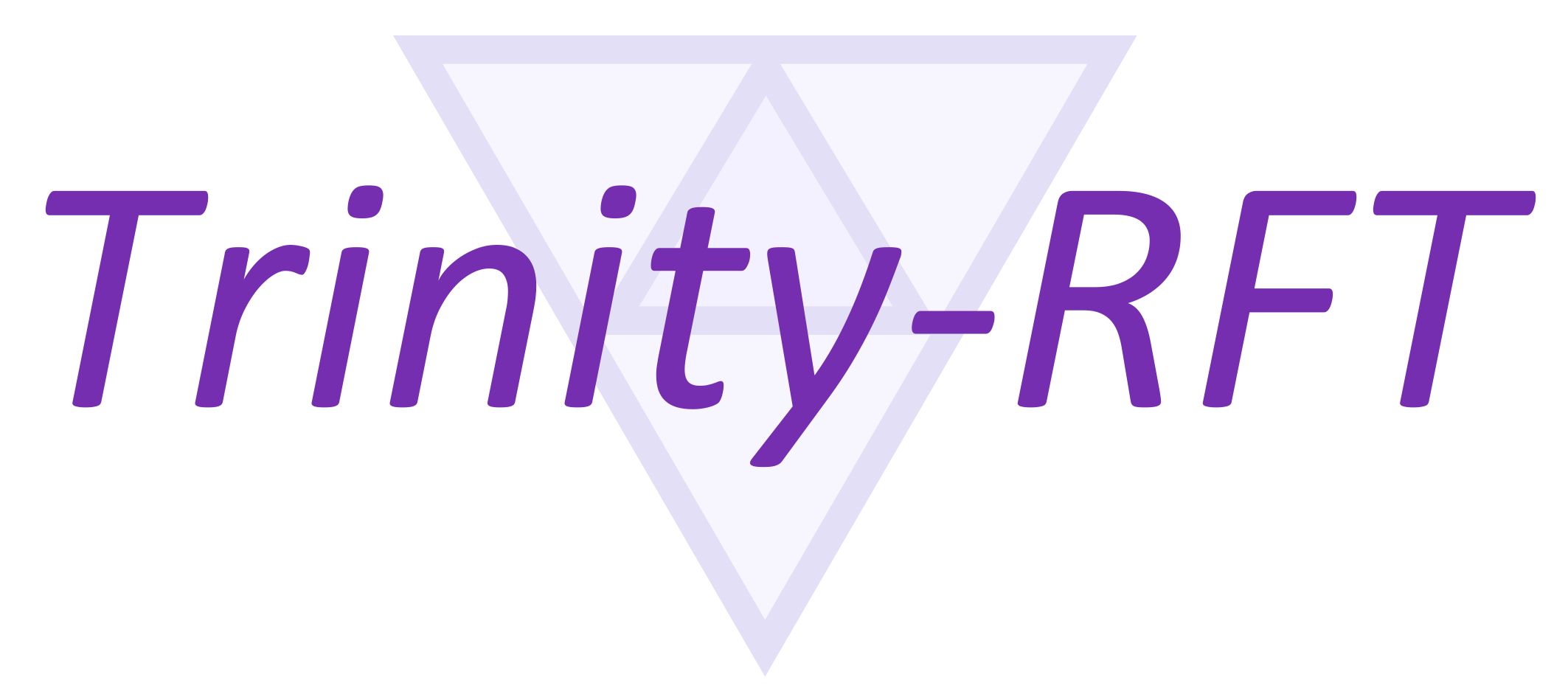}}: 
  A General-Purpose and Unified
  Framework
  for\\
  Reinforcement Fine-Tuning of Large Language Models
}

\author{
  Xuchen Pan$^*$$^{\dagger}$,
  Yanxi Chen$^*$$^{\dagger}$,
  Yushuo Chen$^*$,
  Yuchang Sun$^*$,
  Daoyuan Chen$^*$$^{\dagger}$,\\
  Wenhao Zhang,
  Yuexiang Xie,
  Yilun Huang,
  Yilei Zhang,
  Dawei Gao,
  Weijie Shi,\\
  Yaliang Li$^{\dagger}$,
  Bolin Ding$^{\dagger}$,
  Jingren Zhou\\
  \ \\
  Alibaba Group
}

\date{}

\maketitle

\renewcommand*{\thefootnote}{\fnsymbol{footnote}}
\footnotetext[1]{Equal contribution.}
\footnotetext[2]{Corresponding author. \texttt{\{chenyanxi.cyx,panxuchen.pxc,daoyuanchen.cdy,yaliang.li,bolin.ding\}@alibaba-inc.com}}
\renewcommand*{\thefootnote}{\arabic{footnote}}

\begin{abstract}

\trinity is a general-purpose, unified and easy-to-use framework designed for reinforcement fine-tuning (RFT) of large language models.
It is built with a modular and decoupled design, consisting of 
(1)~an RFT-core that unifies and generalizes synchronous/asynchronous, on-policy/off-policy, and online/offline modes of RFT;
(2)~seamless integration for agent-environment interaction with high efficiency and robustness;
and (3)~systematic data pipelines optimized for RFT.
\trinity can be easily adapted for diverse application scenarios, 
and serves as a unified platform for development and research of advanced reinforcement learning paradigms at both macroscopic and microscopic levels.
This technical report outlines the vision, features, design and implementations of \trinity,
accompanied by extensive examples, applications and experiments that demonstrate its functionalities and user-friendliness.

\end{abstract}

\vspace{1.0em}

\noindent\textbf{GitHub:}
\url{https://github.com/modelscope/Trinity-RFT}

\vspace{0.5em}

\noindent\textbf{Documentation:}
\url{https://modelscope.github.io/Trinity-RFT}

\vspace{0.5em}

\noindent\textbf{Note:}
\trinity is currently under active development.
This technical report corresponds to commit id \texttt{63d4920} (July 14, 2025) of the GitHub repository,
and will be continuously updated as the codebase evolves.
Comments, suggestions and contributions are welcome!

\vspace{1.0em}

\section{Introduction}

Reinforcement learning (RL) has achieved remarkable success in the development of large language models (LLMs).
Examples include aligning LLMs with human preferences via reinforcement learning from human feedback (RLHF) \cite{Ouyang2022},
and training long-CoT reasoning models via RL with rule-based or verifiable rewards (RLVR) \cite{deepseekai2025deepseekr1incentivizingreasoningcapability,kimiteam2025kimik15scalingreinforcement}.
However, such approaches are limited in their abilities to handle dynamic, agentic and continuous learning in the real world.

We envision a future where AI agents learn by interacting directly with environments, collecting lagged or complex reward signals, and continuously refining their behavior through RL based on the collected experiences \cite{Silver2025}.
For example, imagine an AI scientist who designs an experiment, executes it, waits for feedback (while working on other tasks concurrently), and iteratively updates itself based on true environmental rewards and feedback when the experiment is finally finished.

\begin{figure}
  \centering
  \includegraphics[width=1.0\textwidth]{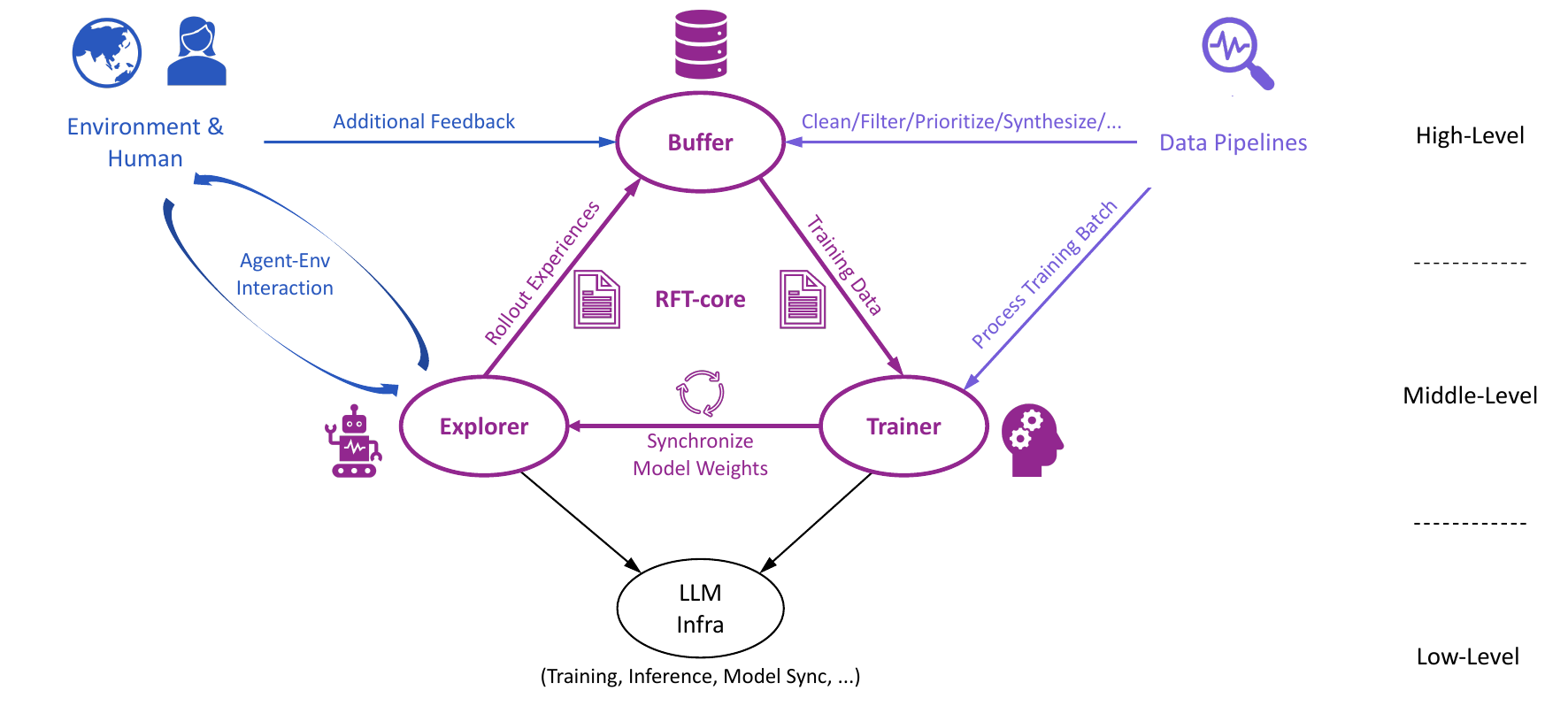}
  \caption{The high-level design of \trinity.}
  \label{fig:trinity_design}
\end{figure}

This vision motivates us to develop \trinity, 
a reinforcement fine-tuning (RFT) framework that aims to offer a path into this future.
The modular, decoupled and trinity design of \trinity illustrated in Figure~\ref{fig:trinity_design},
along with its various features,
makes it a promising solution for realizing such a vision.

\subsection{Key Features of \trinity}
\label{sec:feature}

\trinity is a general-purpose, unified, scalable and user-friendly RL framework
that can be easily adapted for diverse experimental or real-world scenarios.
It integrates both \emph{macroscopic and microscopic} RL methodologies in one place;
roughly speaking, the former deals with natural language and plain text, 
while the latter handles \texttt{torch.Tensor} (such as token probabilities, gradients, and model weights of LLMs)\footnote{Many prior RL works for games/control/LLMs focus mainly on the microscopic aspect, e.g., designing policy loss functions or optimization techniques for updating the policy model.
On the other hand, pre-trained LLMs, as \emph{generative} models with \emph{rich prior knowledge} of natural language and the world,
open up numerous opportunities at the macroscopic level,
e.g., experience synthesis by reflection or reasoning with environmental feedback \cite{da2025agentrlvrtrainingsoftwareengineering}, leveraging existing text processing methods like deduplication and quality filtering \cite{datajuicer}, among others.}.
The key features of \trinity are presented below,
which will be further elaborated in Section~\ref{sec:design_implementation} and exemplified in Section~\ref{sec:examples_experiments}.

\paragraph{An RFT-core that unifies and generalizes diverse RL modes.}

\trinity implements diverse RL methodologies in a unified manner,  
supporting \emph{synchronous/asynchronous, on-policy/off-policy}, and \emph{online/offline} training.
These RL modes can be flexibly generalized,
e.g., a hybrid mode that incorporates expert trajectories to accelerate an online RL process \cite{nachum2017bridginggapvaluepolicy,yan2025learningreasonoffpolicyguidance}.
This unification is made possible partly by our decoupled design (which will soon be introduced in Section~\ref{subsec:rft_core}) that allows rollout and training to be executed separately and scaled up independently on different devices,
while having access to the same stand-alone experience buffer.
The efficacy of various RL modes has been validated empirically by our experiments in Section~\ref{subsec:exp_RL_modes},
which particularly highlight the efficiency gains by off-policy or asynchronous methods.

\paragraph{Agent-environment interaction as a first-class citizen.}

\trinity allows delayed rewards and environmental feedback in multi-step/time-lagged feedback loop, 
handles long-tailed latencies and the straggler effect via asynchronous and streaming LLM inference,
and deals with environment/agent failures gracefully via dedicated timeout/retry/skip mechanisms. 
These together ensure efficiency and robustness of continuous agent-environment interaction in complex real-world scenarios.

\paragraph{Systematic data pipelines optimized for RFT.}

\begin{figure}
    \centering
    \includegraphics[width=\textwidth]{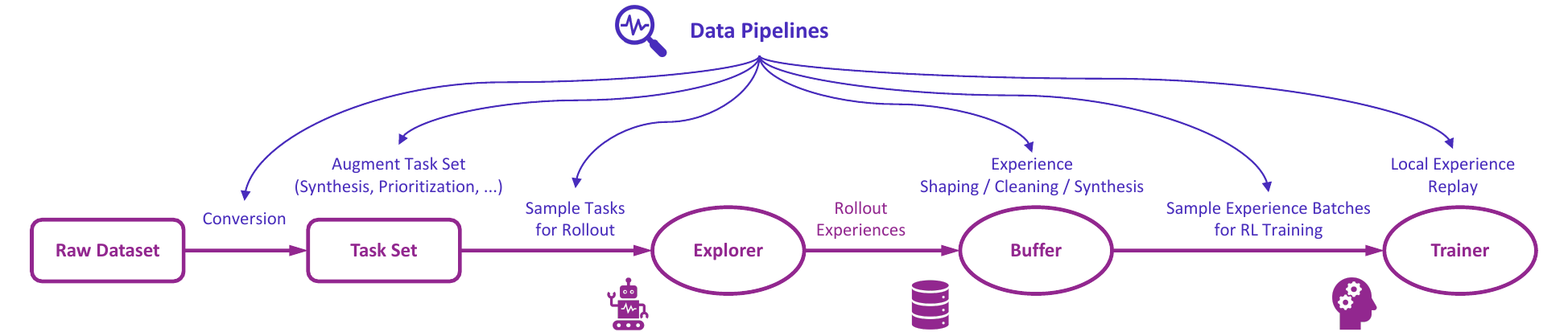}
    \caption{The high-level design of data pipelines in \trinity.
    }
    \label{fig:data_pipelines}
\end{figure}

Figure~\ref{fig:data_pipelines} illustrates the high-level design of data pipelines in \trinity,
which regard rollout tasks and experiences as \emph{dynamic assets} to be actively managed throughout the RFT lifecycle.
\trinity empowers users to: 
(1) \emph{curate tasks for curriculum learning}, e.g., by prioritizing easier tasks at the beginning of training to stabilize and accelerate the learning process;
(2) \emph{actively manipulate experience} by cleaning, filtering, or synthesizing new experiences, such as repairing failed trajectories or amplifying successful ones;
(3) \emph{perform online reward shaping} by augmenting sparse environmental rewards with dense, computed signals, such as quality or diversity scores;
(4) customize interfaces for \emph{human-in-the-loop curation} and utilize an \emph{agentic paradigm for RFT data processing} that translates high-level natural language commands (e.g., ``improve response diversity and safety for coding scenarios'') into complex data pipelines, powered by established community tools like Data-Juicer \cite{datajuicer}.
For instance, 
Section~\ref{subsec:examples_data_processor} presents experiments that demonstrate the efficacy of task prioritization and reward shaping empowered by \trinity.

\paragraph{User-friendliness as a top priority.}

For development and research, the modular and decoupled design of \trinity allows the user to develop new RFT methodologies by adding one or a few small, plug-and-play classes (modified from built-in templates) that implement the essential functionalities of interest, with minimal code duplication or intrusive changes to the codebase.
An example can be found in Section~\ref{subsec:example_mix_algorithm}, 
which shows that three compact python classes (with around 200 lines of code in total) suffice for implementing a hybrid RL process that leverages samples from multiple data sources and updates the policy model with a customized loss function.
For applications, the user can adapt \trinity to a new scenario by simply implementing a single \texttt{Workflow} class that specifies the logic of agent-environment interaction,
{as will be exemplified in Section~\ref{subsec:customizing_agent_env}.}
To further enhance usability, \trinity incorporates various graphical user interfaces to support low-code usage and development, 
enhance transparency of the RFT process, 
and facilitate easy monitoring and tracking.

\subsection{Related Works}

There exist numerous open-source RLHF frameworks, such as 
veRL~\cite{verl}, OpenRLHF~\cite{openrlhf}, TRL~\cite{vonwerra2022trl}, ChatLearn~\cite{chatlearn}, Asynchronous RLHF~\cite{noukhovitch2025faster}, among others.
Some of them have been further adapted for training long-CoT reasoning models or for agentic RL more recently.

Concurrent to \trinity, 
some recent works on LLM reinforcement learning also advocate a decoupled and/or asynchronous design;
examples include StreamRL~\cite{zhong2025streamrlscalableheterogeneouselastic}, MiMo~\cite{coreteam2025mimounlockingreasoningpotential}, AReaL~\cite{fu2025areal}, ROLL~\cite{roll2025alibaba}, LlamaRL~ \cite{wu2025llamarldistributedasynchronousreinforcement}, Magistral~\cite{mistralai2025magistral}, AsyncFlow \cite{han2025asyncflowasynchronousstreamingrl}, among others.

Complementary to this huge number of related works,
\trinity provides the community with a new solution that is powerful, easy-to-use, and unique in certain aspects.
In a nutshell,
\trinity aims to be general-purpose and applicable to diverse application scenarios,
while unifying various RFT modes, RFT methodologies at macroscopic and microscopic levels, 
and RFT-core/agent-environment interaction/data pipelines.
Such a system-engineering perspective makes \trinity particularly useful for \emph{handling the whole RFT pipeline in one place}.
We also hope that some specific features of \trinity, 
such as data persistence in the experience buffer, and distributed deployment of multiple independent explorers,
will open up new opportunities for LLM reinforcement fine-tuning.

\section{Design and Implementations}
\label{sec:design_implementation}

The overall design of \trinity exhibits a trinity
consisting of (1)~\emph{RFT-core}, 
(2)~\emph{agent-environment interaction}, 
and (3)~\emph{data pipelines},
which are illustrated in Figure~\ref{fig:trinity_design} and elaborated in this section.

\subsection{RFT-Core}
\label{subsec:rft_core}

\begin{figure}
  \centering
  \includegraphics[width=.9\textwidth]{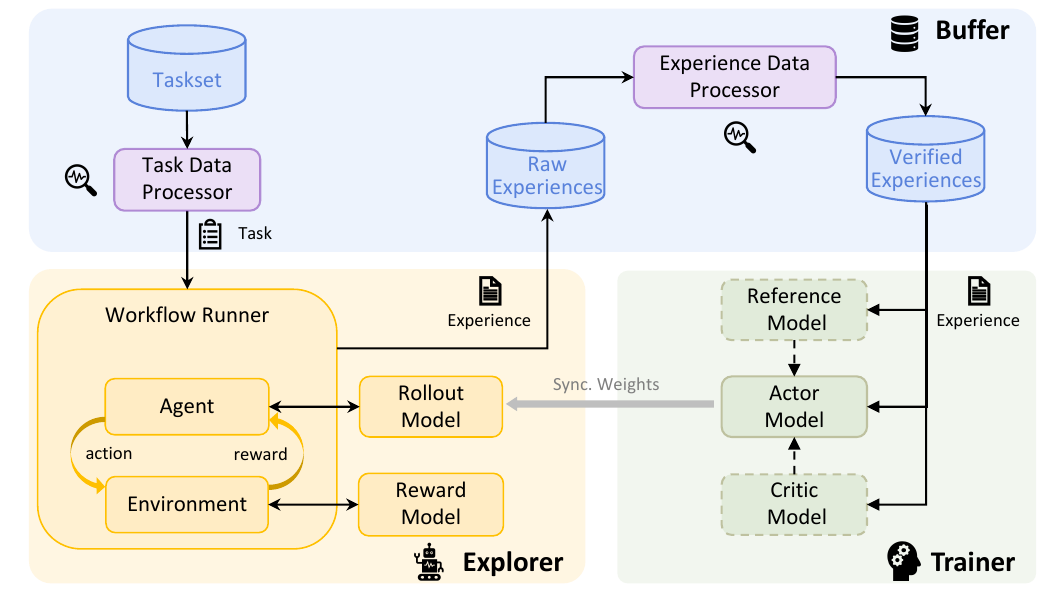}
  \caption{The architecture of RFT-core in \trinity.}
  \label{fig:trinity_architecture}
\end{figure}

RFT-core is the component of \trinity, highlighted at the center of Figure~\ref{fig:trinity_design}, where the core RFT process happens.
Its design also exhibits a trinity, 
consisting of the \emph{explorer}, \emph{buffer}, and \emph{trainer}.
\begin{itemize}
\item The explorer, powered by a rollout model, 
takes a \emph{task} as input and solves it by
executing a \emph{workflow} that specifies the logic of agent-environment interaction,
thereby collecting \emph{experiences} (including rollout trajectories, rewards, and other useful information) to be stored in the buffer.

\item The buffer stores experiences that can be generated by the explorer or by other sources, such as human experts.
It can be realized in various forms, such as a non-persistent \texttt{ray.Queue} or a persistent SQLite database.
It also assists with fetching training samples for the trainer, and can be integrated with advanced sampling strategies and post-processing operations.

\item The trainer, backed by a policy model, samples batches of experiences from the buffer and updates the policy model via RL algorithms.
\end{itemize}
Our implementations allow the explorer and trainer to be deployed on separate machines and act independently.
They are only connected via 
(1)~access to the same experience buffer with a customizable sampling strategy,
and (2)~model weight synchronization by a customizable schedule.
See Figure~\ref{fig:trinity_architecture} for an illustration.
This decoupled design of RFT-core offers support for diverse RFT modes with great flexibility.

\subsubsection{Unified Support for Diverse RFT Modes}
\label{subsubsec:rft_modes}

\begin{figure}
\centering
\includegraphics[width=\textwidth]{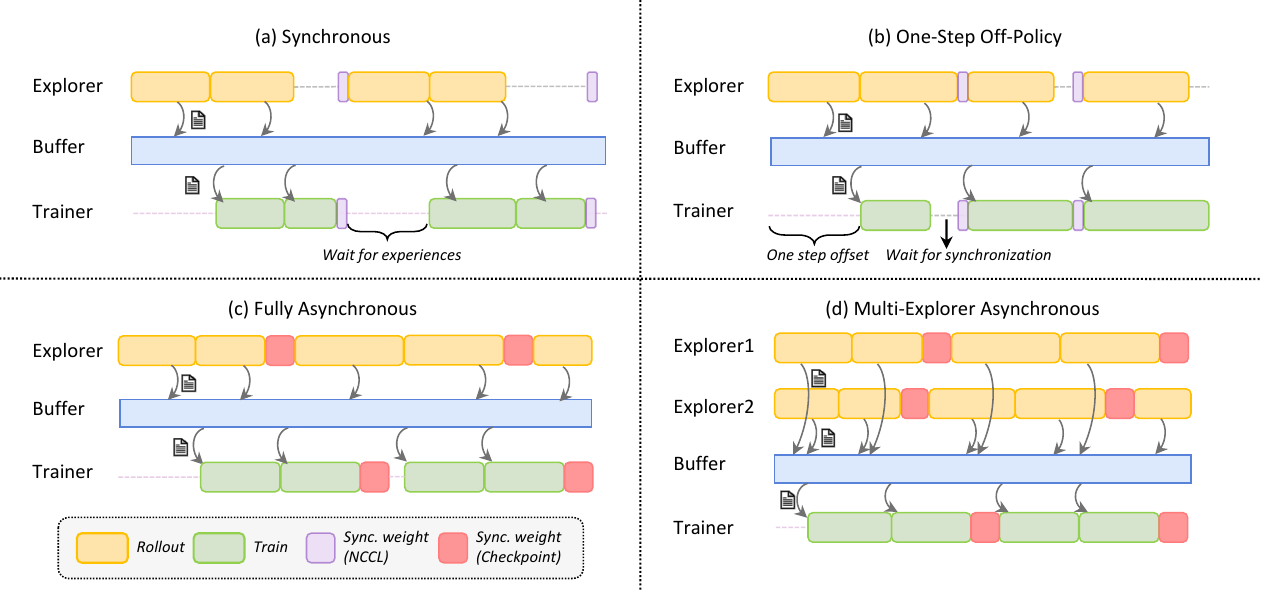}
\caption{A visualization of diverse RFT modes supported by \trinity, including: 
(a) synchronous mode, with \texttt{sync\_interval=2}; 
(b) one-step off-policy mode, with \texttt{sync\_interval=1} and \texttt{sync\_offset=1};
(c) fully asynchronous mode, with \texttt{sync\_interval=2}; 
(d) multi-explorer asynchronous mode, with \texttt{sync\_interval=2}.
The buffer supports, in principle, arbitrary management and sampling strategies for experiences.
}
\label{fig:viz_rl_modes}
\end{figure}

We present the RFT modes supported by \trinity, 
some of which are demonstrated in Figure~\ref{fig:viz_rl_modes}.

\paragraph{Synchronous mode.}

In the {synchronous mode} shown in Figure~\ref{fig:viz_rl_modes} (a),
the explorer and trainer get launched simultaneously,
work in close coordination, 
and synchronize their model weights once every \texttt{sync\_interval} training steps.
Within each synchronization period, the explorer continuously generates \texttt{sync\_interval} batches of rollout experiences and stores them in the buffer,
which are then retrieved and utilized by the trainer for updating the policy model.
If \texttt{sync\_interval=1}, this is a strictly \textbf{on-policy} RL process,
whereas if \texttt{sync\_interval>1}, it becomes \textbf{off-policy} (akin to the mode adopted in \cite{kimiteam2025kimik15scalingreinforcement}) and can be accelerated by pipeline parallelism between the explorer and trainer.
This mode can be activated by setting the configuration parameter \texttt{mode=both}.

\paragraph{One-step off-policy mode.}

This mode, demonstrated in Figure~\ref{fig:viz_rl_modes} (b), 
closely resembles the synchronous mode,
except for an offset of one batch between the explorer and trainer.
This allows the trainer to sample experiences from the buffer immediately after model weight synchronization,
thereby streamlining the execution of explorer and trainer with smaller pipeline bubbles,
at the cost of slight off-policyness.
The visualization in Figure~\ref{fig:viz_rl_modes} (b) corresponds to configuration parameters \texttt{sync\_interval=1} and \texttt{sync\_offset=1}, both of which can take more general values in \trinity.

\paragraph{Asynchronous mode.}

In the fully {asynchronous mode} shown in Figure~\ref{fig:viz_rl_modes} (c),
the explorer and trainer act almost independently.
The explorer continuously generates rollout experiences and stores them in the buffer,
while the trainer continuously samples experiences from the buffer and uses them for training the policy model.
External experiences, e.g., those generated by expert models or humans, can be continuously incorporated into the buffer as well.
The explorer and trainer independently load or save model weights from the checkpoint directory every \texttt{sync\_interval} steps,
keeping the distribution of rollout experiences up to date.
This mode can be activated by setting \texttt{mode=explore/train} and launching the explorer and trainer separately on different GPUs.

\paragraph{Multi-explorer asynchronous mode.}

One benefit brought by the decoupled design is that
explorers and trainers can scale up independently on separate devices.
As a proof-of-concept,
\trinity offers support for a multi-explorer asynchronous mode, as demonstrated in Figure~\ref{fig:viz_rl_modes} (d),
where multiple explores send the generated rollout experiences to the same buffer.
Scaling up the number of independent and distributed explorers can be particularly useful for resolving \emph{data scarcity} and speeding up the generation of experiences in real-world scenarios where rollout trajectories have to be sampled via interaction with the physical world, or in an environment with sparse and lagged feedback.
Another by-product of this multi-explorer mode is \emph{24-hour non-interrupted service} for real-world online serving situations:
since the explorers can pause and update model weights at different moments,
it can be guaranteed that there is always one explorer ready to serve an incoming request immediately whenever it arrives.
This is in contrast to a single-explorer mode, where online service has to be paused when the explorer is updating its model weights.

\paragraph{Benchmark mode.}

\trinity supports a {benchmark mode} that allows the user to evaluate one or multiple checkpoints on arbitrary benchmarks,
after the RFT training process has finished.
To activate this mode, the user simply needs to set \texttt{mode=bench} 
and specify the paths for the evaluation datasets in the configurations.
This mode can be particularly useful for experimental purposes;
for example,
the user might want to try out different RFT techniques or configurations quickly (with limited evaluation on hold-out data) during training,
identify which RFT trials have achieved stable convergence and high rewards,
and then conduct more thorough evaluations only for the checkpoints of these successful trials.

\paragraph{Train-only mode.}

In certain scenarios, the user would like to train the policy model without further exploration,
using experiences that have already been collected and stored in the buffer.
This {train-only mode} can be activated by setting the configuration parameter \texttt{mode} $=$ \texttt{train} and launching the trainer alone.
Offline methods like Supervised Fine-Tuning (SFT) and Direct Preference Optimization (DPO) \cite{dpo} can be regarded as special cases of such scenarios, 
both of which are natively supported by \trinity.
For another example, consider an online RFT process that expands over a long period,
where the explorer alone is launched during the daytime for serving human users and collecting experiences,
while the trainer alone is launched at night for updating the policy model 
(which will be thoroughly validated and evaluated before it can be actually deployed as the rollout model for the next day).

\paragraph{Discussions.}

We conclude this subsection with two remarks.
(1)~Given the unified implementation of various RFT modes, 
it is easy to design and implement a \emph{hybrid mode} with \trinity that combines multiple modes into a single learning process.
One example is learning with both online rollout data and offline-collected expert data,
via jointly optimizing two loss terms corresponding to these two data sources.
Section~\ref{subsec:example_mix_algorithm} illustrates how to implement this conveniently in \trinity.
(2)~We take a \emph{system-algorithm co-design} perspective in the development of \trinity,
aiming to unify and generalize diverse RFT methodologies in this framework.
RFT-core provides the necessary infrastructure for achieving this goal.
This technical report focuses on the system perspective,
and we refer interested readers to the literature for recent algorithmic developments in off-policy / asynchronous RL for LLMs \cite{nachum2017bridginggapvaluepolicy,dong2023raft,richemond2024offlineregularisedreinforcementlearning,kimiteam2025kimik15scalingreinforcement,fletberliac2025contrastivepolicygradientaligning,noukhovitch2025faster,wang2025distrl,xu2025traininglargelanguagemodels,yan2025learningreasonoffpolicyguidance,yao2025grouprelativereinforce}.

\subsubsection{Implementations of RFT-Core}

We present some implementation details of RFT-core in the following.

\paragraph{Inference and training engines.}

The current version of \trinity leverages vLLM \cite{vllm} as the inference engine for the explorer,
which offers features including paged attention, continuous batching \cite{orca}, asynchronous and concurrent inference for multiple rollout trajectories, among others.
\trinity also leverages verl~\cite{verl} as the training engine for the trainer,
which gracefully handles model placement (for the policy, critic and reference models) and incorporates various performance optimizations for training (such as dynamic batching, management of padding and unpadding, etc.).
\trinity stands on the shoulders of these excellent open-source projects, and will continue to benefit from their future development.

\paragraph{Experience buffer.}

\trinity supports multiple types of experience buffers,
ranging from a non-persistent \texttt{ray.Queue} to persistent SQLite or Redis databases.
While using a basic first-in-first-out queue is the most straightforward approach, 
data persistence with a database opens up many new opportunities (e.g., advanced sampling strategies), 
as discussed throughout this report.
\trinity has provided dedicated read/write control to prevent any conflict in accessing the buffer.

\paragraph{Model weight synchronization.}

\trinity supports model weight synchronization between the explorer and trainer by NCCL \cite{nccl}, 
or by checkpoint saving and loading.
The former is faster (when available),
while the latter is generally more flexible and widely applicable, especially for asynchronous RFT modes.

\subsection{Agent-Environment Interaction}

To adapt \trinity to a new downstream scenario,
the user mainly needs to define and register a customized workflow 
(by inheriting the base class \texttt{Workflow} or \texttt{MultiTurnWorkflow})
where the logic of agent-environment interaction for this particular scenario is implemented.
Advanced methods for experience synthesis with environmental feedback \cite{da2025agentrlvrtrainingsoftwareengineering} can be implemented in the same way as well.
See Section~\ref{subsec:customizing_agent_env} for detailed examples.
The workflow will then be executed by workflow runners within the explorer for generating experiences,
as shown in Figure~\ref{fig:trinity_architecture}.

Numerous challenges arise when one tries to build an RFT framework that can efficiently and robustly handle
real-world interaction between the LLM-powered agent and the environment.
These include long-tailed latencies, agent/environment failures, and lagged reward signals, among others.
\trinity regards agent-environment interaction as a first-class citizen
and incorporates various solutions to tackle these challenges, for example:
\begin{itemize}
    \item The workflow runners in \trinity support \emph{asynchronous and streaming generation} of rollout trajectories for multiple tasks. This helps mitigate the straggler effect caused by the long-tailed latencies in rollout generation and agent-environment interaction, thereby accelerating the RFT process. 
    \emph{Load balancing} among multiple LLM inference engines within one RFT training course is also taken care of, and would be one direction for further optimizing the utilization of computational resources.
    \item \trinity incorporates various \emph{timeout/retry/skip mechanisms} for fault tolerance and robustness, which ensure that continuous rollout generation would not be interrupted or blocked by individual failures in certain rounds of agent-environment interaction. 
    This is crucial for stable and efficient learning in real-world scenarios, e.g., when the agent interacts with a large number of MCP services \cite{mcp} that differ vastly in quality and availability.
    \item \trinity is built to provide native support for \emph{asynchronous RFT modes}, which allow great flexibility in the paces of the explorer and trainer. This can boost the overall efficiency of the RFT process, compared to synchronous modes where the slower one among the explorer and trainer can block the progress of the other and cause waste of computational resources. 
    \item For \emph{lagged reward signals}, the trinity design of RFT-core offers a natural solution. 
    As soon as the rollout trajectory (without reward values) has been generated, it is saved into the experience buffer, but marked as ``not ready for training''. 
    The explorer is now free from this task and may continue to collect experiences for other tasks.
    When the reward signals from the environment finally arrive, they are written to the buffer, and the corresponding experience is now marked as ``ready for training''.
    \item For multi-turn conversations and ReAct-style workflows \cite{yao2023react}, \trinity supports concatenating multiple rounds of agent-environment interaction compactly into a single sequence, with proper masking that indicates which tokens need to be incorporated into the training objective of RL algorithms.
    This avoids unnecessary recomputation and thus improves training efficiency,
    compared to a vanilla approach that represents a $K$-turn rollout trajectory with $K$ separate samples.
    \item As another performance optimization,
    the implementation of \trinity allows resetting the environment in a workflow, rather than re-initializing it every time.
    This is especially useful for scenarios where setting up the environment is costly.

\end{itemize}

\subsection{Data Pipelines}
\label{sec:data-pipelines}

The data pipelines in \trinity aim to address fundamental challenges in RFT scenarios,
such as managing {heterogeneous data dynamics} across interaction workflows, enabling {delayed reward integration}, and facilitating {continuous data curation}. 
Our solutions center on four core aspects: \emph{end-to-end data transformation}, \emph{task curation}, \emph{active experience shaping}, and \emph{human-in-the-loop curation}, each corresponding to key requirements identified in our development of RFT-core (Section~\ref{subsec:rft_core}).
\subsubsection{End-to-end Data Transformation}
\label{sec:data-pipelines-transform}

\begin{figure}
    \centering
    \includegraphics[width=\textwidth]{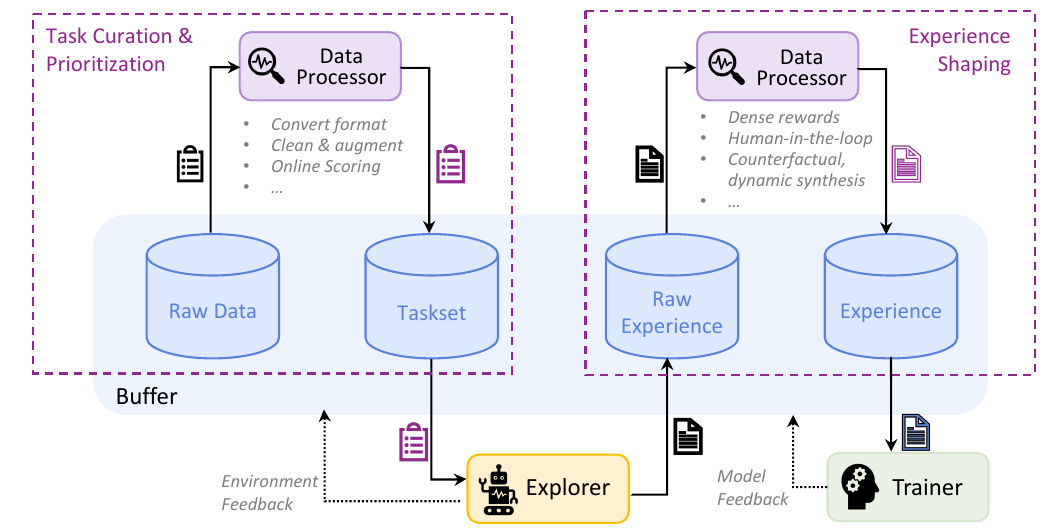}
    \caption{The interaction of data processor and data buffers in \trinity, divided into two key stages. \textbf{Left}: Task Curation \& Prioritization prepares the initial tasks for the explorer. \textbf{Right}: Experience Shaping processes the collected trajectories from the explorer before they are used by the trainer. The data processor is a central component that operates on different buffers at different stages.
    }
    \label{fig:data_pipeline_buffers}
\end{figure}

To support the diverse RFT modes (e.g., synchronous or asynchronous) in \trinity,
we establish a service-oriented data pipeline architecture as illustrated in Figure~\ref{fig:data_pipeline_buffers}. It decouples data pipeline logic from procedure control to enable flexible RL-oriented data transformations with two key modules:

\begin{itemize}
    \item 
The \textbf{Formatter Module} unifies disparate data sources into RFT-compatible formats, providing convenient conversion between raw inputs (e.g., meta-prompts, domain-specific corpora, and QA pairs with tagged rewards) and structured RFT representations.
For efficient RFT workloads, we utilize buffer-based persistent storage (Section \ref{subsec:rft_core}) to support different data models, such as \texttt{ExperienceModel} for prioritized rollout trajectories and \texttt{DPODataModel} for preference pairs.
The conversion logic and data models are highly customizable to meet diverse requirements for managing experience data. This flexibility enables robust metadata recording and field normalization, which is essential for advanced scenarios such as asynchronous RFT in trainer-explorer environments, agent self-evolution from a cold start using meta-prompts, and knowledge injection from structurally complex domain-specific corpora.

\item The \textbf{Controller Module} manages the complete data pipeline lifecycle through \emph{distributed server initialization}, \emph{declarative configuration}, and \emph{automated dataset persistence}. It implements dynamic control mechanisms for asynchronous scenarios and protection against resource exhaustion, with configurable termination conditions based on compute quota or data quantity. 
This modular design enables \trinity to handle data transformations flexibly while maintaining consistency across different RFT modes. 
\end{itemize} 

The {Formatter-Controller} duality mirrors the explorer-trainer decoupling in RFT-core, enabling parallel data ingestion and model updating. This design also allows \trinity to handle delayed rewards through version-controlled experience updates while maintaining low-latency sampling for the trainer.

\subsubsection{Task Curation and Prioritization}
\label{sec:data-pipelines-curation}
Before the RFT loop begins, it is crucial to prepare a high-quality set of initial tasks. This stage, depicted on the left side of Figure~\ref{fig:data_pipeline_buffers}, transforms raw data into an optimized task set for the explorer. 

The process begins with raw data sources (e.g., prompts, domain corpora), which are ingested into a buffer. The \textbf{Data Processor}, powered by over 100 operators from Data-Juicer~\cite{datajuicer}, reads from this buffer to perform various curation tasks. It provides composable building blocks for experience cleaning (e.g., length filters, duplication removal), safety alignment (e.g., toxicity detection, ethics checks), and preference data synthesis (e.g., critique-conditioned augmentation).
By treating Data-Juicer as a modular data processing operator pool rather than a central dependency, \trinity provides RL-specific abstractions and coherence, while benefiting from well-established data tools.

The processed data is then organized into a structured task buffer. This stage effectively implements a form of \textbf{curriculum learning} by allowing users to prioritize tasks (e.g., from easy to hard), guiding the explorer towards a more efficient and stable learning trajectory from the outset. This entire workflow is managed by a service-oriented architecture that decouples data logic from procedural control, ensuring flexibility and scalability, especially in asynchronous and distributed settings.

\subsubsection{Active Experience Shaping}
\label{sec:data-pipelines-shape}
Once the explorer begins interacting with the environment, it generates a continuous stream of experience data. To maximize learning efficiency, this raw experience must be actively shaped before it reaches the trainer. This stage is shown on the right side of Figure~\ref{fig:data_pipeline_buffers}.

Generated experiences are first collected in a buffer. The \textbf{Data Processor} is applied again with a series of transformations to \emph{clean, augment, or synthesize} these experiences. This is where the core of RFT data intelligence lies. Key capabilities include:
\begin{itemize}
    \item \textbf{Agent-Driven Data Processing}: 
    \trinity introduces a powerful agentic paradigm for data manipulation. Users can define complex processing pipelines through high-level objectives, specified as either \textit{natural language commands} (e.g., ``improve safety'' or ``increase response diversity'') or explicit Data-Juicer configurations. The framework automatically translates these commands into executable workflows backed by its modular components like \texttt{DataCleaner} and \texttt{DataSynthesizer}. 
    This design provides a user-friendly abstraction layer over the underlying Data-Juicer operators, making advanced processing functionalities accessible to both RFT users familiar with Data-Juicer and those who are not.
    It also facilitates the flexible injection of user-defined inductive biases into the learning process, unlocking new research directions for self-evolving agents, as we will discuss later in Section \ref{sec:data-pipelines-future}.

    \item \textbf{Online Reward Shaping}: The data processor can dynamically augment the reward signal. Instead of relying on a single, often sparse, task-completion reward, users can add dense rewards based on quality, diversity, or safety scores computed on the fly. This enriched feedback provides a much stronger learning signal for the trainer.

    \item \textbf{Prioritized Experience Replay}: Experiences are not treated equally. \trinity allows for flexible, multi-dimensional utility scoring to prioritize the most valuable samples for training. The \texttt{DataActiveIterator} supports version-controlled experience reuse and cross-task data lineage tracking, ensuring that the trainer always learns from the most informative data available. This mechanism is also critical for handling delayed rewards, as experience utilities can be updated asynchronously as new feedback arrives.
\end{itemize}

\subsubsection{Human-AI Collaboration}
\label{sec:data-pipelines-human}

In scenarios where human feedback is irreplaceable,
\trinity establishes a bi-directional human-AI collaboration loop that provides first-class support for human annotations, 
based on Label Studio \cite{LabelStudio} and Data-Juicer's HumanOPs.

\begin{itemize}
    \item \textbf{Multi-stage annotation.} \trinity implements configurable procedures combining automatic pre-screening and human verification. Typical stages include preference annotation (comparative assessment of model responses), quality auditing (human verification of automated cleaning/synthesis results), and cold-start bootstrapping (initial dataset curation through expert demonstrations).
    \item \textbf{Native asynchronism support.} As the collection of human feedback is generally slower than AI/model feedback, we provide dedicated capabilities to handle both synchronous and asynchronous feedback modes, with configurable timeout and polling parameters. The feedback collaboration is based on an event-driven design, with automatic task creation upon data state changes, configurable notifications via email/Slack/webhook, and an atomic transaction model for annotation batches.
    \item \textbf{Customization.}
    Different applications may involve humans in heterogeneous ways. 
    We thus prioritize flexibility in both the interaction-interface and service levels.
    Examples include
    rich built-in interfaces that can be extended in a visualized style with XML-like tags provided by Label Studio, 
    fine-grained quality scoring for reward shaping, free-form feedback attachment for dataset shaping, among others. 
    Moreover, for easy deployment, we provide local Label Studio instance management with automatic environment setup via Docker/pip; optimized SDK interactions with batch request coalescing; unified logging across annotation tools and ML services; and concurrent annotation campaigns through priority-based task routing, while maintaining full data lineage preserved via \texttt{LineageTracker}. 
\end{itemize}
The decoupled design of \trinity, and the presence of a standalone experience buffer in particular, 
enable {human feedback} to participate in RL loops without breaking the \emph{asynchronous execution model}. For instance, human-verified samples can be prioritized for training while fresh experiences are being collected, which is a critical capability for real-world deployment scenarios with mixed feedback sources. Further details for human-AI collaboration in \trinity will be illustrated in Section~\ref{sec:exapme:human-loop}.

\subsubsection{Discussion: Unlocking New Research \& Development Directions}
\label{sec:data-pipelines-future}

The modular design of our data pipelines and the powerful data processor 
open up promising research and development avenues to be further explored.

One direction is about \emph{effective management of experience data}. While prior RFT works often treat the experience as a static log, \trinity enables a more sophisticated, full-lifecycle approach to data, from selective acquisition to efficient representation:
\begin{itemize}
    \item \emph{Intelligent Perception and Collection:} In an open-ended environment, what experience is ``worth'' recording? Storing everything creates a low signal-to-noise ratio and burdens the trainer. \trinity's architecture allows researchers to implement \emph{active collection strategies}. For instance, one could design a data processor operator that evaluates incoming experiences from the explorer based on metrics like surprise, uncertainty, or information gain, and only commits the most salient trajectories to the replay buffer. This transforms data collection from passive logging into a targeted, intelligent process.
    
    \item \emph{Adaptive Representation:} Raw experience is often high-dimensional and redundant (e.g., long dialogues, complex code generation traces). How can this be distilled into a format that an agent can efficiently learn from? The data processor in \trinity acts as a powerful transformation engine. Researchers can use it to explore various representation learning techniques, such as automatically summarizing trajectories, extracting causal chains from tool usage, or converting a multi-turn dialogue into a structured preference pair. This not only makes training more efficient but also opens the door to building meta-experience (more abstract and reusable knowledge) from raw interaction data.

    \item \emph{Agentic Workflows:} \trinity's agent-driven processing enables the research and development of \textit{self-improving agents}, e.g., by configuring the policy agent to also serve as the ``processing agent'' for LLM-based Data-Juicer operators. Such an agent could perform its own critique and dynamically curate its own training data, creating a truly autonomous learning and data management loop.
\end{itemize}

Another direction is about \emph{synthetic and counterfactual experience processing}. The integration of synthesis operators enables research into creating ``better-than-real'' data. Instead of relying solely on the agent's own trial-and-error, our framework facilitates exploring questions like:
\begin{itemize}
    \item \emph{Dynamic and Composable Rewarding:} With our framework, researchers can move beyond static, hand-crafted rewards. It is now possible to investigate \emph{dynamic reward shaping}, where auxiliary signals like novelty, complexity, or alignment scores are automatically extracted from trajectories and composed into a dense reward function. How to define ``good'' experience and how can we learn the optimal combination of these reward components as the agent's policy evolves?
    \item \emph{Experience Reorganization:} Can successful sub-trajectories from different tasks be ``spliced'' together to solve a novel, composite task? For example, can an agent that has learned to ``open a door'' and ``pick up a cup'' synthesize a new trajectory to "enter the room and fetch the cup"?
    \item \emph{Failure Repair:} Can the data processor identify where errors occur in a failed trajectory, and synthesize a corrected version for the trainer to learn from, effectively turning failures into valuable lessons?
    \item \emph{Success Amplification:} Can a single successful experience be augmented into multiple diverse yet successful variants, thereby improving the generalization and robustness of the learned policy?
\end{itemize}

By providing dedicated capabilities for such advanced data and reward manipulation, \trinity aims to facilitates flexible processing of ``experience data'' for the next generation of self-evolving LLMs.

\subsection{User-Friendliness}

\trinity has been designed with user-friendliness as a top priority.

\paragraph{For development and research:}

\begin{figure}
    \centering
    \includegraphics[width=0.95\textwidth]{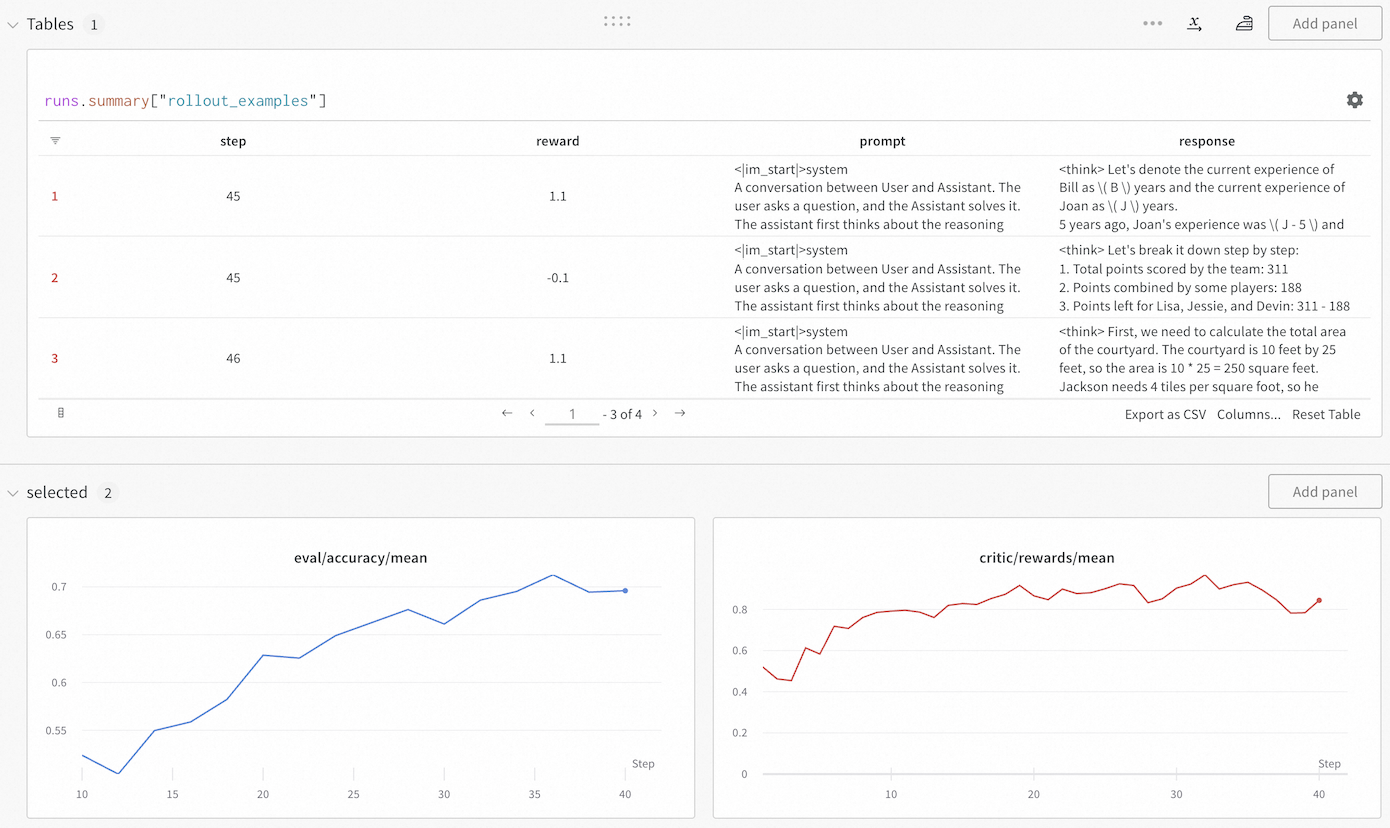}
    \caption{A snapshot of the monitor implemented in \trinity.
    }
    \label{fig:monitor_example}
\end{figure}

The modular and decoupled design of \trinity allows users to develop a new algorithm for a specific aspect of RFT by adding one or a few new classes that implement the essential functionalities of interest, 
without concerning about other aspects of RFT or intrusive modifications of the original codebase.
In addition,
we include a monitor (built upon Wandb~\cite{wandb} and TensorBoard~\cite{tensorboard}) that makes it easy to track the progress of an RFT process, both quantitatively (e.g., via learning curves for rewards and other metrics) and qualitatively (e.g., via concrete examples of rollout trajectories generated at different RL steps). 
See Figure~\ref{fig:monitor_example} for an example snapshot of the monitor.

\paragraph{For RFT applications:}

\begin{figure}
    \centering

    \begin{subfigure}{.45\textwidth}
        \includegraphics[width=\textwidth]{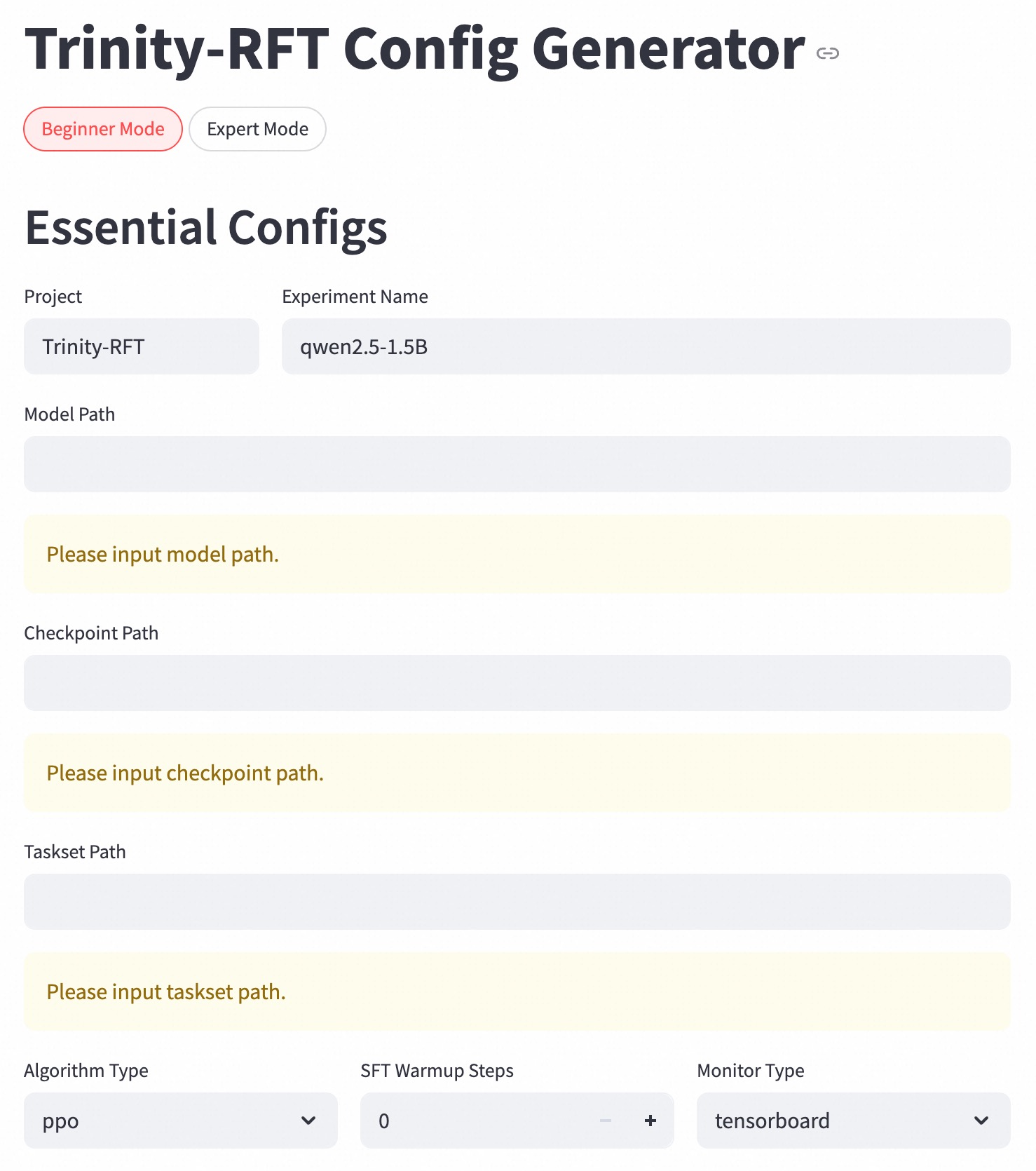}
        \caption{The ``beginner'' mode.}
    \end{subfigure}%
    \begin{subfigure}{.45\textwidth}
        \includegraphics[width=\textwidth]{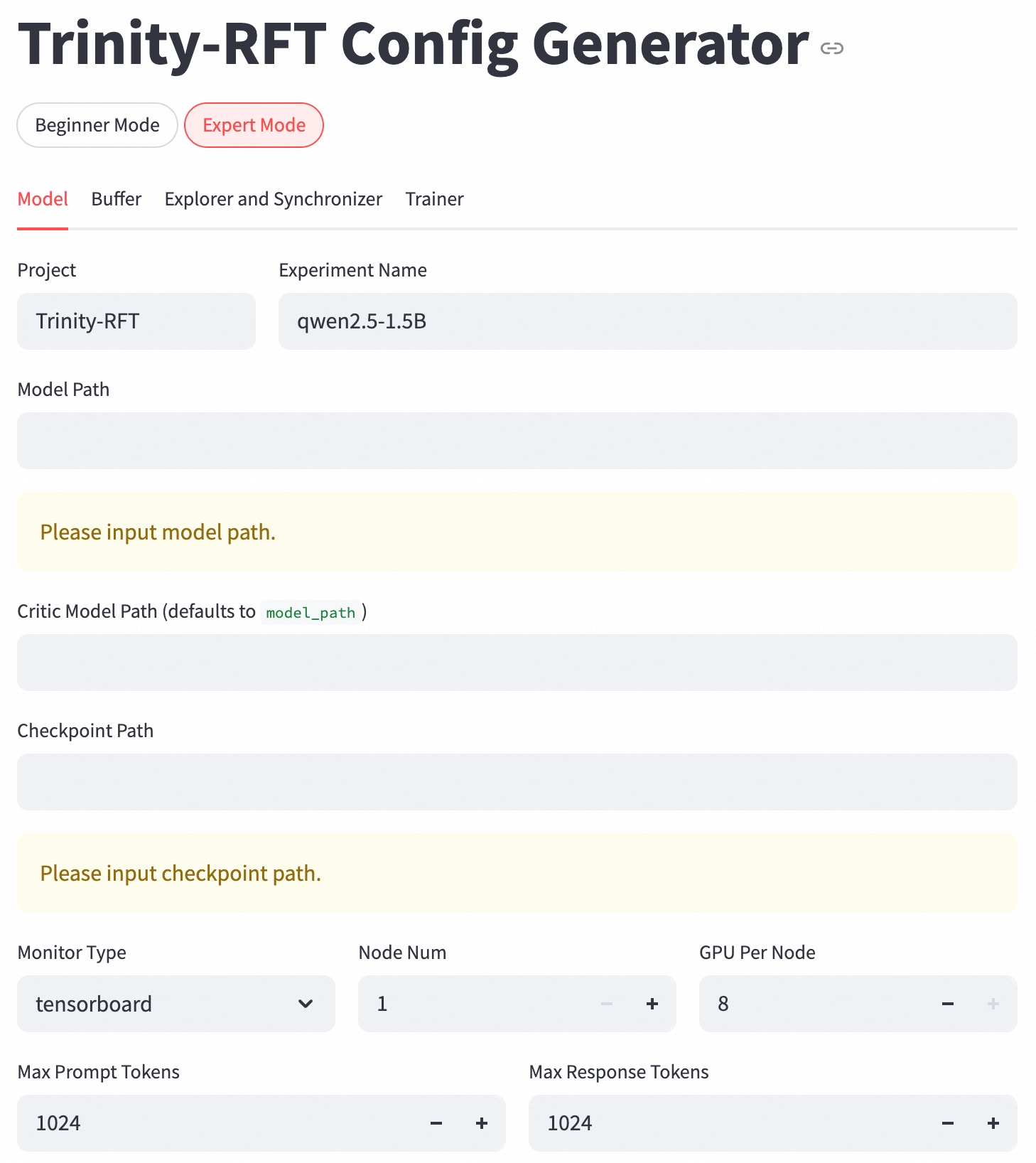}
        \caption{The ``expert'' mode.}
    \end{subfigure}
    \caption{Snapshots of the configuration manager.
    }
    \label{fig:config_manager}
\end{figure}

\trinity offers extensive graphical user interfaces to support low-code usage of the framework,
and to maximize transparency of the RFT process.
For example, we implement a configuration manager, as shown in Figure~\ref{fig:config_manager}, that allows the user to create configuration files conveniently via a front-end interface.
We also provide \trinitystudio, an all-in-one unified UI (including the aforementioned monitor and configuration manager) that allows the user to configure and run data inspection, data processing, RFT learning process, etc., all by clicking the mouse and filling forms, without writing any code.
An example for using \trinitystudio will be introduced in Section \ref{subsec:trinity-studio}.
Such functionalities, of course, can be useful not only for applications but also for development and research.

\section{Examples, Applications, and Experiments}
\label{sec:examples_experiments}

This section demonstrates the utilities and user-friendliness of \trinity and exemplifies some concepts introduced in previous sections, 
through a diverse range of examples, applications and experiments.
Additional step-by-step tutorials can be found on the documentation website\footnote{\url{https://modelscope.github.io/Trinity-RFT}},
or the \texttt{examples} folder of the GitHub repository\footnote{\url{https://github.com/modelscope/Trinity-RFT/tree/main/examples}}.

\subsection{Customizing Agent-Environment Interaction}
\label{subsec:customizing_agent_env}

With a modular design, \trinity can be easily adapted to a new downstream scenario by implementing the logic of agent-environment interaction in a single workflow class, without modifications to other components of the codebase.
This approach is also sufficient for macroscopic RL algorithm design that targets high-quality experience synthesis with environmental feedback \cite{da2025agentrlvrtrainingsoftwareengineering}.
We provide some concrete examples in the rest of this subsection.

\subsubsection{Single-turn Scenarios}

In a simple yet common scenario, 
a user of \trinity would like to train an LLM for completing single-turn tasks,
where the LLM generates one response to each input query.
For this purpose, the user mainly needs to 
(1)~define and register a single-turn workflow class (by inheriting the base class \texttt{Workflow}) tailored to the targeted tasks, and
(2)~specify the tasksets (for training and/or evaluation) and the initial LLM, both of which are compatible with HuggingFace \cite{huggingface} and ModelScope \cite{modelscope} formats.

Listing~\ref{lst:example_basic_workflow} gives a minimal example for implementing a single-turn workflow.
Suppose that each task is specified by a \texttt{<question, answer>} tuple.
The \texttt{run()} method of \texttt{ExampleWorkflow} calls the LLM once to generate a response for the question, 
calculates its reward,
and returns an \texttt{Experience} instance that consists of the response itself, the reward value, and the log-probabilities of response tokens predicted by the rollout model (which is necessary for certain RL algorithms, such as PPO \cite{schulman2017proximalpolicyoptimizationalgorithms} and GRPO \cite{shao2024deepseekmathpushinglimitsmathematical}).
Some built-in workflows and reward functions have been implemented in \trinity, e.g., the \texttt{MathWorkflow} class for math-related tasks.

In some cases, the user wants to utilize auxiliary LLMs in the workflow, e.g., for computing rewards via LLM-as-a-judge,
or for playing the roles of other agents in a multi-agent scenario.
For these purposes, the user can specify \texttt{auxiliary\_models} via APIs when initializing the workflow.

\begin{lstlisting}[language=Python, label={lst:example_basic_workflow}, caption=A minimal example for implementing a customized workflow.]
@WORKFLOWS.register_module("example_workflow")
class ExampleWorkflow(Workflow):

    def __init__(
        self,
        model: ModelWrapper,
        task: Task,
        auxiliary_models: Optional[List[openai.OpenAI]] = None,
    ):
        super().__init__(model, task, auxiliary_models)
        self.question = task.raw_task.get("question")
        self.answer = task.raw_task.get("answer")

    def calculate_reward_by_rule(self, response: str, truth: str) -> float:
        return 1.0 if response == truth else 0.0

    def calculate_reward_by_llm_judge(self, response: str, truth:str) -> float:
        judge_model = self.auxiliary_models[0]
        PROMPT_FOR_JUDGE = """Please evaluate..."""
        completion = judge_model.chat.completions.create(
            model="gpt-4",  # Or another suitable judge model
            messages=[{"role": "user", "content": PROMPT_FOR_JUDGE}],
        )
        reward_str = completion.choices[0].message.content.strip()
        reward = float(reward_str)
        return reward

    def run(self) -> List[Experience]:
        response = self.model.chat(
            [
                {
                    "role": "user",
                    "content": f"Question:\n{self.question}",
                }
            ],
            **self.rollout_args,
        )
        reward: float = self.calculate_reward_by_rule(response.response_text, self.answer)
        # reward: float = self.calculate_reward_by_llm_judge(response.response_text, self.answer)
        return [
            Experience(
                tokens=response.tokens,
                prompt_length=response.prompt_length,
                reward=reward,
                logprobs=response.logprobs,
            )
        ]
\end{lstlisting}

\subsubsection{Multi-turn Scenarios}

In more advanced cases, 
the user would like to train an LLM-powered agent that solves multi-turn tasks by repeatedly interacting with the environment.
In \trinity, achieving this is mostly as simple as in the single-turn case,
except that the user needs to define and register a multi-turn workflow class
by inheriting the base class \texttt{MultiTurnWorkflow}.
Listing~\ref{lst:example_multi_turn_workflow} provides one such example using the ALFWorld dataset \cite{shridhar2021alfworld}.
For training efficiency, the \texttt{process\_messages\_to\_experience()} method concatenates multiple rounds of agent-environment interactions compactly into an \texttt{Experience} instance consisting of a single token sequence with proper masking,
which can readily be consumed by standard RL algorithms like PPO and GRPO.

For more detailed examples of multi-turn cases, please refer to the documentation\footnote{\url{https://modelscope.github.io/Trinity-RFT/tutorial/example_multi_turn.html}}.

\begin{lstlisting}[language=Python, label={lst:example_multi_turn_workflow}, caption=An implementation of a multi-turn workflow for ALFWorld \cite{shridhar2021alfworld}.]
@WORKFLOWS.register_module("alfworld_workflow")
class AlfworldWorkflow(MultiTurnWorkflow):
    """A workflow for the ALFWorld task."""

    def generate_env_inference_samples(self, env, rollout_num) -> List[Experience]:
        print("Generating env inference samples...")
        experience_list = []
        for i in range(rollout_num):
            observation, info = env.reset()
            final_reward = -0.1
            memory = []
            memory.append({"role": "system", "content": AlfWORLD_SYSTEM_PROMPT})
            for r in range(self.max_env_steps):
                format_obs = format_observation(observation)
                memory = memory + [{"role": "user", "content": format_obs}]
                response_text = self.model.chat(memory, n=1)[0].response_text
                memory.append({"role": "assistant", "content": response_text})
                action = parse_action(response_text)
                observation, reward, done, info = env.step(action)
                if done:
                    final_reward = reward
                    break
            experience = self.process_messages_to_experience(
                memory, final_reward, {"env_rounds": r, "env_done": 1 if done else 0}
            )
            experience_list.append(experience)
        # Close the env to save CPU memory
        env.close()
        return experience_list

    def run(self) -> List[Experience]:
        # ...
        game_file_path = self.task_desc
        rollout_n = self.repeat_times
        # ...
        env = create_environment(game_file_path)
        return self.generate_env_inference_samples(env, rollout_n)


\end{lstlisting}

\subsubsection{Experience Synthesis in Workflows}

As mentioned in Section~\ref{sec:feature},
\trinity has been designed to streamline RL algorithm design and development at both macroscopic and microscopic levels.
One example for the former is \emph{experience synthesis}:
at each RL step, 
the agent (backed by the rollout LLM) iteratively generates refined responses to a query by incorporating feedback or guidance from the environment, 
which can be in the form of plain text rather than numerical reward values.
The resulting data will then be utilized for updating the policy model, e.g., by a standard SFT or RL loss.
Such a macroscopic RL approach is made possible by pre-trained LLMs' generative nature and rich prior knowledge about natural language.
Closely related to this idea is Agent-RLVR \cite{da2025agentrlvrtrainingsoftwareengineering}, 
a contemporary work that applies such an approach to software engineering scenarios.

Within \trinity,
this process of experience synthesis can be regarded as a particular way of agent-environment interaction,
and thus can be realized by simply implementing a \texttt{Workflow} class.
As a minimal demonstration, suppose that we want to implement this approach for a math reasoning scenario,
where the agent generates multiple rollout responses to an input query, 
receives feedback from the environment regarding correctness of the responses,
reflects on the gathered information,
and generates a final response to the query. 
Listing~\ref{lst:example_exp_synthesis_workflow} presents an implementation of this approach within \trinity.

\begin{lstlisting}[language=Python, label={lst:example_exp_synthesis_workflow}, caption=A toy implementation of experience synthesis with environmental feedback.]
@WORKFLOWS.register_module("reflect_once_workflow")
class ReflectOnceWorkflow(Workflow):

    def run(self) -> List[Experience]:
        experiences = []

        # Stage 1: K-rollout generation
        rollout_messages = self.create_rollout_messages()
        responses = self.model.chat(
            rollout_messages,
            n=self.k_rollouts,
            temperature=self.temperature,
            logprobs=self.logprobs,
            max_tokens= self.task.rollout_args.max_tokens,
        )
        rollout_responses = [response.response_text.strip() for response in responses]

        # Stage 2: Verification
        verification_results = []
        for rollout_response in rollout_responses:
            is_correct = self.verify_answer(rollout_response, self.ground_truth)
            verification_results.append(is_correct)

        # Stage 3: Reflection
        reflection_messages = self.create_reflection_messages(
            rollout_responses, 
            verification_results,
        )
        reflection_responses = self.model.chat(
            reflection_messages,
            n=1,
            temperature=self.temperature,
            logprobs=self.logprobs,
            max_tokens=self.task.rollout_args.max_tokens,
        )
        reflection_response = reflection_responses[0]

        # Verify the reflection response
        reflection_text = reflection_response.response_text.strip()
        reflection_is_correct = self.verify_answer(reflection_text, self.ground_truth)

        if reflection_is_correct:
            sharegpt_message = [
                {
                    "role": "system",
                    "content": self.task.format_args.system_prompt
                },
                {
                    "role": "user",
                    "content": self.question
                },
                {
                    "role": "assistant",
                    "content": reflection_text
                }
            ]
            experience = self.process_messages_to_experience(sharegpt_message)
            experiences.append(experience)

        # Save experience to file
        if self.exp_file and sharegpt_message is not None:
            exp_data = sharegpt_message
            self.exp_file.write(json.dumps(exp_data, ensure_ascii=False) + "\n")
            self.exp_file.flush()
        return experiences

\end{lstlisting}

\subsection{RL Algorithm Development with \trinity}
\label{subsec:example_mix_algorithm}

To support RL algorithm development,
\trinity allows researchers and developers to focus on designing and implementing the essential logic of a new RL algorithm,
without the need to care about the internal engineering details about \trinity.

As an example, suppose that we want to implement a \mix algorithm that seamlessly integrates online RL and offline SFT into a single learning process.
In its most basic form, the \mix algorithm requires that 
(1)~the trainer samples from two sources of experiences,
i.e., the rollout experiences collected online and the high-quality expert trajectories collected offline;
and (2)~the trainer updates its policy model with a loss function that handles both sources of experiences properly,
e.g., a weighted sum of GRPO loss for the on-policy rollout experiences and SFT loss for the expert trajectories.

Variants of this \mix algorithm include 
adaptive weighting of multiple loss terms \cite{fu2025srftsinglestagemethodsupervised},
alternating between RL and SFT \cite{ma2025learningreinforcementlearningcant},
incorporating expert trajectories into RL loss \cite{nachum2017bridginggapvaluepolicy,song2023hybrid,yan2025learningreasonoffpolicyguidance},
or incorporating SFT loss for high-reward rollout trajectories generated by older versions of the rollout model \cite{expreplay}.
Such approaches have proved to be effective in accelerating the online RL process with only a small amount of expert experiences,
or to enhance stability and plasticity in continual learning.

\begin{figure}
\centering
\includegraphics[width=.8\textwidth]{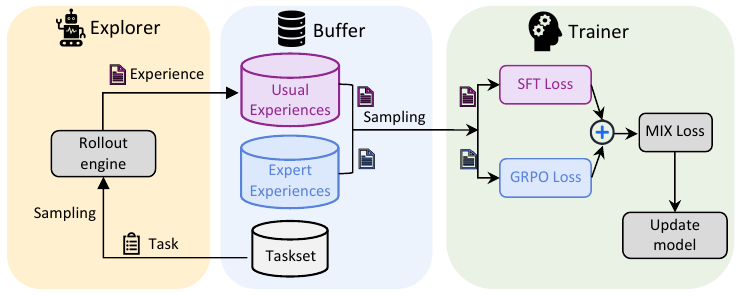}
\caption{A visualization of the \mix algorithm.}
\label{fig:viz_mix_algo}
\end{figure}

The \mix algorithm is visualized in Figure~\ref{fig:viz_mix_algo}, where we integrate GRPO loss for usual experiences generated by the rollout model and SFT loss for expert experiences into a unified training pipeline. 
It requires dealing with different sources of experiences, and two types of loss functions;
fortunately, to implement such an algorithm in \trinity, we only need to define three new classes --- \texttt{MixSampleStrategy}, \texttt{MIXPolicyLossFn}, and \texttt{MIXAlgorithm} --- as demonstrated in Listing~\ref{lst:example_mix_algorithm}.
With these components, \trinity enables a seamless integration of online RL and offline SFT within the same training loop.
More details of the \mix algorithm are referred to the documentation\footnote{\url{https://modelscope.github.io/Trinity-RFT/tutorial/example_mix_algo.html}}.

\begin{lstlisting}[language=Python, label={lst:example_mix_algorithm}, caption=An implementation of the \mix algorithm with \trinity.]
@SAMPLE_STRATEGY.register_module("mix")
class MixSampleStrategy(SampleStrategy):
    def __init__(self, buffer_config: BufferConfig, trainer_type: str, **kwargs):
        # ...
        self.usual_exp_buffer = get_buffer_reader(
            buffer_config.trainer_input.experience_buffer, usual_buffer_config
        )
        self.expert_exp_buffer = get_buffer_reader(
            buffer_config.trainer_input.sft_warmup_dataset, expert_buffer_config
        )
        # ...
    
    def sample(self, step: int) -> Tuple[Any, Dict, List]:
        """Sample a batch composed of rollout experiences and expert trajectories"""
        usual_exp_list = self.usual_exp_buffer.read()
        expert_exp_list = self.expert_exp_buffer.read()
        exp_list = usual_exp_list + expert_exp_list
        exps = Experiences.gather_experiences(exp_list, self.pad_token_id)
        # ...


@POLICY_LOSS_FN.register_module("mix")
class MIXPolicyLossFn(PolicyLossFn):
    def __init__(self, mu: float = 0.1, ...):
        # ...
        self.mu = mu
        self.grpo_loss_fn = PPOPolicyLossFn(...)
        self.sft_loss_fn = SFTLossFn(...)

    def __call__(
        self,
        logprob: torch.Tensor,
        old_logprob: torch.Tensor,
        action_mask: torch.Tensor,
        advantages: torch.Tensor,
        is_expert_mask: torch.Tensor,
        **kwargs,
    ) -> Tuple[torch.Tensor, Dict]:
        """Calculate a weighted sum of GRPO loss and SFT loss"""
        # ...
        grpo_loss, grpo_metrics = self.grpo_loss_fn(
            logprob[~is_expert_mask],
            old_logprob[~is_expert_mask],
            action_mask[~is_expert_mask],
            advantages[~is_expert_mask],
            **kwargs,
        )
        sft_loss, sft_metrics = self.sft_loss_fn(
            logprob[is_expert_mask],
            action_mask[is_expert_mask],
        )
        loss = (1 - self.mu) * grpo_loss + self.mu * sft_loss
        # ...
        return loss, metrics

@ALGORITHM_TYPE.register_module("mix")
class MIXAlgorithm(AlgorithmType):
    """MIX algorithm."""

    use_critic: bool = False
    use_reference: bool = True
    use_advantage: bool = True
    can_balance_batch: bool = True
    schema: type = ExperienceModel

    @classmethod
    def default_config(cls) -> Dict:
        return {
            "repeat_times": 8,
            "policy_loss_fn": "mix",   # Specify the MIX loss function
            "advantage_fn": "grpo",
            "sample_strategy": "mix",  # Specify the MIX sampling strategy
        }

\end{lstlisting}

\subsection{Unified Support for Diverse RL Modes}
\label{subsec:exp_RL_modes}

As explained previously in Section~\ref{subsubsec:rft_modes},
\trinity offers support for synchronous/asynchronous, on-policy/off-policy, and online/offline RL,
controlled by a few configuration parameters.
In this subsection, we conduct experiments for comparing the following RL modes:
\begin{itemize}
    \item The synchronous mode: \texttt{mode=both}, \texttt{sync\_interval=\{1,2,10\}}, \texttt{sync\_offset=0};
    \item The one-step off-policy mode: \texttt{mode=both}, \texttt{sync\_interval=1}, \texttt{sync\_offset=1};
    \item The fully asynchronous mode: the explorer and trainer are launched with \texttt{mode=explore} and \texttt{train} respectively, with \texttt{sync\_interval=10}.
\end{itemize}
Our experiments include \emph{dummy} learning processes (which will soon be explained) for performance profiling,
as well as \emph{real} learning processes with vanilla GRPO \cite{shao2024deepseekmathpushinglimitsmathematical} in different modes.

\subsubsection{Experiments: Performance Profiling}

\paragraph{Settings.}

We aim to measure and compare the efficiency of different RL modes under controlled settings.
It is noteworthy that, even with all other variables controlled, 
different RL modes can still result in different trained models --- and thus different rollout response lengths --- during the learning processes,
which have direct impacts on performance metrics like wall-clock time and GPU utilization rate.

To mitigate this,
we conduct performance profiling with \emph{dummy} learning processes, where the learning rate is set to {zero}.
A dummy learning process closely resembles a real learning process,
in that all necessary computation and communication (e.g., rollout generation, gradient computation, model weight synchronization) are executed;
the only difference is that the rollout model (and thus the distribution of rollout trajectories) remains fixed throughout and same across different RL modes.

To show the performance of \trinity in diverse scenarios, we consider both math reasoning task (GSM8k~\cite{cobbe2021gsm8k}) and multi-turn agentic task (ALFWorld~\cite{shridhar2021alfworld}).
In the experiments, we use the Qwen2.5-Instruct~\cite{qwen25technicalreport} model series of different sizes (1.5B, 3B, and 7B), 
and run the GRPO~\cite{deepseekai2025deepseekr1incentivizingreasoningcapability} algorithm (with 8 rollout trajectories per task) in all modes.
We choose a 100-step training trace to evaluate the performance and report the following metrics: 
(1) \emph{end-to-end wall-clock time} and \emph{time speedup}: the wall-clock time from the start of running the command to the end of finishing 100 training steps; 
(2) \emph{GPU utilization}: the GPU utilization in percent for each GPU; 
(3) \emph{GPU power usage}: the GPU power usage as a percentage of its power capacity for each GPU.
Metrics for GPU utilization and power usage were extracted from WandB\footnote{\url{https://docs.wandb.ai/guides/models/app/settings-page/system-metrics/}} and averaged over all GPUs.
We run each experiment for three random trials and report the mean and standard deviation.
Unless specified otherwise, each experiment uses 8 NVIDIA A100-80G GPUs.

\paragraph{Results for GSM8k.}
We use the 2/6 GPU partition for explorer/trainer. While this configuration is not the optimal one for all experiments, it is sufficient to show the difference between several RL modes.
In our GSM8k experiments, the batch size is 96 tasks, and the temperature is 1.0.
The results for both Qwen2.5-1.5B-Instruct and Qwen2.5-7B-Instruct models are shown in Table~\ref{tab:performance_profiling_gsm8k}.

We observe that in the synchronous mode (with \texttt{sync\_offset=0}), a less frequent synchronization (i.e., a larger \texttt{sync\_interval}) effectively improves the efficiency, GPU utilization, and GPU power usage. 
This is mainly because, as shown in Figure~\ref{fig:viz_rl_modes} (a), 
the impacts of pipeline bubbles in this mode can be effectively reduced by using a lower synchronization frequency.
Besides, Table~\ref{tab:performance_profiling_gsm8k} shows that one-step off-policy and fully asynchronous modes also accelerate the training process with higher GPU utilization, compared to a strictly on-policy mode.
In one-step off-policy mode, the trainer leverages the one-step off-policy data stored in the buffer without needing to wait for new experiences generated by the explorer after weight synchronization, which significantly reduces the GPU idle ratio.
In fully asynchronous mode, the explorer and trainer operate almost independently while fully leveraging GPU resources, except when loading or saving model checkpoints.

\begin{table}
\centering
\caption{Performance profiling for GSM8k with 2/6 GPU partition for explorer/trainer.} 
\label{tab:performance_profiling_gsm8k}
\small
\begin{tabular}{lcccc}
\toprule
\multirow{2}{*}{Mode} & \multicolumn{4}{c}{Qwen2.5-1.5B-Instruct} \\
\cmidrule(lr){2-5}
& Speedup $\uparrow$ & {Time (Minutes)} $\downarrow$ & GPU Utilization (\%) $\uparrow$ & {GPU Power Usage (\%)} $\uparrow$\\
\midrule
Sync. (\texttt{sync\_interval=1}) & $1.00 \times$ & $38.70 \pm 0.34$ & $33.64 \pm 2.15$ & $35.85 \pm 1.83$ \\
Sync. (\texttt{sync\_interval=2}) & $1.24 \times$ & $31.19 \pm 0.08$ & $36.05 \pm 0.49$ & $38.74 \pm1.47$ \\
Sync. (\texttt{sync\_interval=10}) & $1.59 \times$ & $24.28 \pm 0.16$ & $\mathbf{38.27} \pm 0.98$ & $\mathbf{44.41} \pm 0.81$ \\
One-step off-policy & $1.25 \times$ & $30.84 \pm 0.20$ & $32.39\pm 1.17$ & $39.70 \pm 0.78$ \\
Fully async. & $\mathbf{1.61} \times$ & $\mathbf{23.97} \pm 0.03$ & $36.04 \pm 0.61$ & $43.91 \pm 0.48$ \\
\bottomrule
\end{tabular}
\vspace{1em}
\begin{tabular}{lcccc}
\toprule
\multirow{2}{*}{Mode} & \multicolumn{4}{c}{Qwen2.5-7B-Instruct} \\
\cmidrule(lr){2-5}
& Speedup $\uparrow$ & {Time (Minutes)} $\downarrow$ & GPU Utilization (\%) $\uparrow$ & {GPU Power Usage (\%)} $\uparrow$\\
\midrule
Sync. (\texttt{sync\_interval=1}) & $1.00 \times$ & $68.71 \pm 0.54$ & $55.61 \pm 0.80$ & $52.88 \pm 0.29$ \\
Sync. (\texttt{sync\_interval=2}) & $1.31 \times$ & $52.44 \pm 0.41$ & $64.88 \pm 1.35$ & $61.90 \pm 1.32$ \\
Sync. (\texttt{sync\_interval=10}) & $\mathbf{1.85} \times$ & $\mathbf{37.17} \pm 0.15$ & $\mathbf{78.44} \pm 1.03$ & $\mathbf{77.77} \pm 0.96$ \\
One-step off-policy & $1.69 \times$ & $40.73 \pm 0.57$ & $77.19 \pm 2.26$ & $76.17 \pm 1.56$ \\
Fully async. & $1.63 \times$ & $42.17 \pm 1.06$ & $73.90 \pm 2.00$ & $72.74 \pm 1.82$ \\
\bottomrule
\end{tabular}
\end{table}

\begin{table}
\centering
\caption{Performance profiling for ALFWorld with 4/4 GPU partition for explorer/trainer.}
\label{tab:performance_profiling_alfworld}
\small
\begin{tabular}{lcccc}
\toprule
\multirow{2}{*}{Mode} & \multicolumn{4}{c}{Batch\_Size = 4} \\
\cmidrule(lr){2-5}
& Speedup $\uparrow$ & {Time (Minutes)} $\downarrow$ & GPU Utilization (\%) $\uparrow$ & {GPU Power Usage (\%)} $\uparrow$\\
\midrule
Sync. (\texttt{sync\_interval=1}) & $1.00 \times$ & $ 333.68 \pm 0.06$   & $17.19 \pm 0.58$ & $28.44 \pm 0.37$  \\
Sync. (\texttt{sync\_interval=2}) & $1.70 \times$  & $196.64 \pm 0.59$   & $21.69 \pm 0.18$ & $31.35 \pm 0.06$ \\
Sync. (\texttt{sync\_interval=10}) & $5.21 \times$  & $64.09 \pm 0.39$ & $32.85 \pm 0.18$  & $40.86 \pm 0.58$  \\
One-step off-policy & $0.98 \times$  & $340.12 \pm 3.99$   & $14.63 \pm 1.17$ &  $28.21 \pm 0.48$  \\
Fully async. & $\mathbf{5.45} \times$  & $\mathbf{61.27} \pm 0.35$ & $\mathbf{36.46} \pm 0.10$  & $ \textbf{42.51} \pm 0.72$  \\
\bottomrule
\end{tabular}
\vspace{1em}
\begin{tabular}{lcccc}
\toprule
\multirow{2}{*}{Mode} & \multicolumn{4}{c}{Batch\_Size = 32} \\
\cmidrule(lr){2-5}
& Speedup $\uparrow$ & {Time (Minutes)} $\downarrow$ & GPU Utilization (\%) $\uparrow$ & {GPU Power Usage (\%)} $\uparrow$\\
\midrule
Sync. (\texttt{sync\_interval=1}) & $1.00 \times$ & $ 561.21 \pm 2.04$   & $39.37 \pm 0.89$ & $39.93 \pm 0.22$  \\
Sync. (\texttt{sync\_interval=2}) & $1.13 \times$  & $496.80 \pm 0.36$   &37.74 $ \pm 0.39$ & $41.90 \pm 0.44$ \\
Sync. (\texttt{sync\_interval=10}) & $1.59 \times$  & $352.11 \pm 0.49$ & $44.50 \pm 0.58$  & $49.95 \pm 0.61$  \\
One-step off-policy & $1.14 \times$  & $494.13 \pm 0.28$   & $34.89 \pm 0.75$ &  $43.05 \pm 0.81$  \\
Fully async. & $\mathbf{1.65} \times$  & $\mathbf{339.51} \pm 0.24$ & $\mathbf{45.55} \pm 0.20$  & $ \textbf{50.77} \pm 0.45$  \\
\bottomrule
\end{tabular}%
\end{table}

\paragraph{Results for ALFWorld.}
A particular feature of ALFWorld is the long-horizon multi-turn interaction with the environment.
To accommodate the heavy computational demands in rollout, we use the 4/4 GPU partition for explorer/trainer.
In our ALFWorld experiments, we use the Qwen2.5-3B-Instruct model and set the rollout temperature to 1.0.

The results with different batch sizes are shown in Table~\ref{tab:performance_profiling_alfworld}.
One observation is that, when the batch size is 4 tasks, the one-step off-policy mode exhibits no efficiency advantage over the synchronous mode with \texttt{sync\_interval=1}.
This phenomenon can be attributed to the computational imbalance between the explorer and trainer.
In ALFWorld, the larger computation latency of the explorer emerges primarily from 
(1) the inherent complexity of multi-turn environment interactions, and 
(2) the long-tailed latency distribution when certain tasks require extended rollout durations, whose effect is further exacerbated by a small batch size.
The one-step off-policy mode cannot eliminate pipeline bubbles caused by long-tailed latencies in the explorer,
whereas this can be mitigated by the synchronous mode with a large \texttt{sync\_interval} as well as by the asynchronous mode,
thanks to the implementation of streaming rollout generation in \trinity.
Another observation, due to the same reason, is that when scaling the batch size from 4 to 32, all modes incur a small increase (much smaller than $8\times$) in wall-clock time for the same number of training steps, thanks to better GPU usage.

\subsubsection{Experiments: Real Learning with Vanilla GRPO}

\paragraph{Settings.}

We aim to compare the real learning processes by different RL modes.
For simplicity and controlled variability, we use the vanilla GRPO \cite{shao2024deepseekmathpushinglimitsmathematical} algorithm for all RL modes, without specific algorithm design for asynchronous or off-policy cases.
GRPO mainly relies on the mechanism of clipping probability ratio (defaults to the range of $1 \pm 0.2$) to handle the off-policyness of experiences.
For future works, we will investigate more advanced off-policy or asynchronous RL algorithms, %
and develop new ones to accommodate diverse RL modes.
To encourage the exploration of the rollout model, we disable the Kullback-Leibler (KL) penalty or loss in our experiments.

\paragraph{Training.}
For each RL mode, we fine-tune the Qwen2.5-7B-Instruct model for one epoch on the OpenR1-Math-46k~\cite{yan2025learningreasonoffpolicyguidance}\footnote{\url{https://huggingface.co/datasets/Elliott/Openr1-Math-46k-8192}} dataset, a filtered version of the OpenR1-Math-220k\footnote{\url{https://huggingface.co/datasets/open-r1/OpenR1-Math-220k}} dataset.
The allocated GPU ratio for the explorer and trainer is 4/4.
We set the batch size to 120 tasks, the rollout number per task to 32, and the learning rate to 1e-6.

\paragraph{Evaluation.}
For each RL mode, we save the checkpoint of the rollout model once every 100 steps, evaluate the checkpoints using the \texttt{bench} mode, and report the best results among the checkpoints.
The models are evaluated on several math benchmarks, including AIME2024\footnote{\url{https://huggingface.co/datasets/math-ai/aime25}}, AIME2025\footnote{\url{https://huggingface.co/datasets/math-ai/aime25}}, AMC\footnote{\url{https://huggingface.co/datasets/math-ai/amc23}}, and MATH500\footnote{\url{https://huggingface.co/datasets/HuggingFaceH4/MATH-500}}.
For AIME2024, AIME2025, and AMC, we generate 32 responses (with temperature 0.6) per task and report the average accuracy (Avg@32), to ensure the accuracy of the evaluation process; for MATH500, we report Avg@4 as the dataset is relatively large.
For more detailed comparisons, we also plot some training metrics, including reward, response length, gradient norm, and KL distance from the initial LLM, with the wall-clock time as the X-axis.

\paragraph{Results.}

\begin{figure}
\centering
\includegraphics[width=\textwidth]{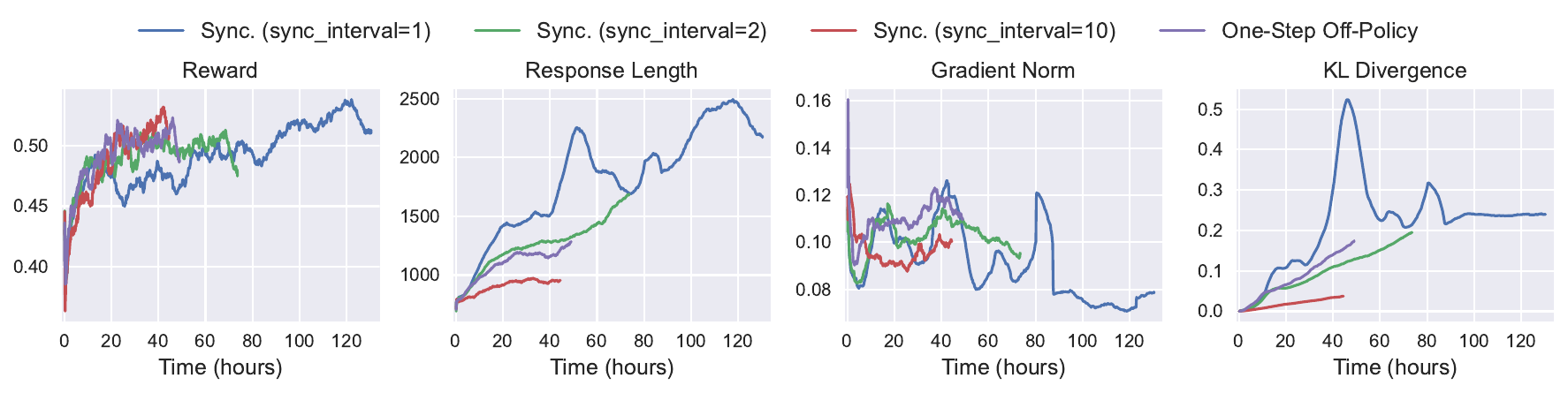}
\caption{Results of training for one epoch by vanilla GRPO in different RL modes. The results are smoothed using a 40-step moving average for clarity.}
\label{fig:exp_real_learning_process}
\end{figure}

Figure~\ref{fig:exp_real_learning_process} presents the training curves. 
It is observed that several RL modes show increasing trends in terms of rewards and response lengths.
The synchronous mode with \texttt{sync\_interval=1} exhibits longer responses and larger KL divergence than other RL modes, 
likely because it updates the rollout model most frequently and leverages on-policy data in each step.

Table~\ref{tab:exp_real_learning_process} presents the evaluation results.
We observe that, for the synchronous mode with \texttt{sync\_offset=0},
increasing \texttt{sync\_interval} reduces the total training time for one epoch, at the cost of slightly compromising the average evaluation performance.
In contrast, the one-step off-policy mode with \texttt{sync\_interval=1} achieves comparable or better performance than the other modes on several benchmarks, 
while achieving around $2.66\times$ speedup in wall-clock time over the strictly on-policy mode.

\begin{table}
\centering
\caption{Performance comparison of different RL modes.}
\label{tab:exp_real_learning_process}
\small
\begin{tabular}{lcccccccc}
\toprule
 & AIME2024 & AIME2025 & AMC & MATH500 & Average & Runtime (Hours) \\
\midrule
Qwen2.5-7B-Instruct & 11.15  & 6.95 & 51.35  & 70.96 & 35.10 & N/A \\
\midrule
Sync. (\texttt{sync\_interval=1}) & 14.58 & \textbf{14.06} & \textbf{61.25} & \textbf{76.25} & \textbf{41.54} & 130.33 \\
Sync. (\texttt{sync\_interval=2}) & 15.52 & 11.67 & 57.97 & 75.15 & 40.08 & 73.57 \\
Sync. (\texttt{sync\_interval=10}) & 14.38 & 12.71 & 57.66 & 75.05 & 39.95 & \textbf{44.43} \\
One-Step Off-Policy & \textbf{16.88} & 12.19 & 59.92 & 74.55 & 40.89 & 48.98 \\
\bottomrule
\end{tabular}
\end{table}

\subsection{Data Processors for Tasks and Experiences}
\label{subsec:examples_data_processor}
We present practical examples to demonstrate how the data pipeline concepts described in Section~\ref{sec:data-pipelines} are applied in \trinity. These use cases highlight how manipulating data at both the \emph{task} and \emph{experience} level directly improves RFT performance and provides granular control over the agent's learning process.

\subsubsection{Static Task Prioritization for Curriculum Learning}
\label{sec:gsm8k-example}

A common strategy for effective training is to present tasks in increasing order of difficulty. This use case demonstrates how \trinity facilitates {curriculum learning} by prioritizing tasks \emph{before} exploration begins. This is particularly crucial for RFT, as it helps stabilize the initial learning phase of the explorer and prevents it from getting stuck on overly complex tasks, leading to a more efficient exploration path.

As shown in Listing~\ref{lst:dataset_config_yaml}, a user can configure this pipeline with a simple YAML file \footnote{The full configuration files can be accessed at \href{https://github.com/modelscope/Trinity-RFT/blob/main/examples/grpo_gsm8k/gsm8k.yaml}{baseline\_run} and \href{https://github.com/modelscope/Trinity-RFT/blob/main/examples/grpo_gsm8k_task_pipeline/gsm8k.yaml}{priority\_run}.}. Here, we use the GSM8K mathematical reasoning dataset. The user provides a natural language instruction via \texttt{dj\_process\_desc}: ``Please compute difficulty scores for these math questions.''. \trinity's data service then orchestrates a three-phase process:
\begin{enumerate}
    \item The data processor invokes an LLM (Qwen-Max) to score the difficulty of each math problem.
    \item The system prioritizes samples with lower difficulty scores, creating an easy-to-hard ordering (by setting `priority\_weights["difficulty"]: -1.0').
    \item The curated and prioritized data is formatted into an RL-ready task set for the explorer.
\end{enumerate}

\begin{figure}
  \centering
  \includegraphics[width=.8\textwidth]{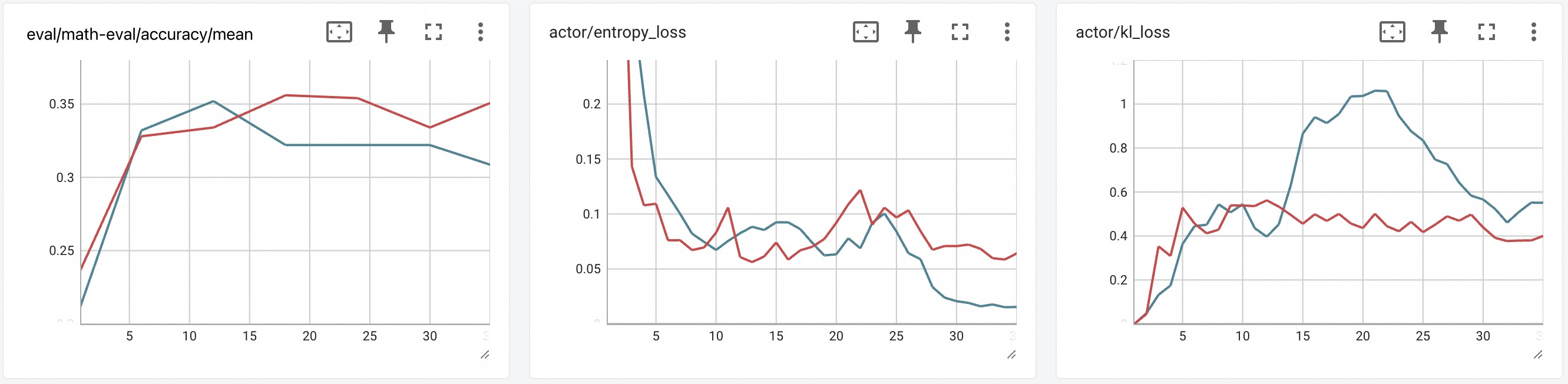}
  \caption{Performance on a math reasoning task. Prioritizing tasks from easy to hard (red line) leads to faster and better convergence compared to the default setting (blue line).}
  \label{fig:data-workflow-static-priority}
\end{figure}

As shown in Figure~\ref{fig:data-workflow-static-priority}, this simple curation strategy yields more stable performance gains. This pattern is highly extensible: users can easily customize the difficulty metric, apply it to their own datasets, or even make the prioritization dynamic by re-ranking tasks periodically based on the agent's current performance.

\begin{lstlisting}[language=YAML, label={lst:dataset_config_yaml}, caption={Data processor configuration, applied on customizable buffers.}]
# Core dataset configuration
data_processor:
  data_workflow_url: "http://127.0.0.1:5005/data_workflow"
  task_pipeline:
    # I/O buffers
    input_buffers:
      - name: "raw_input"
        path: "openai/gsm8k"
        storage_type: "file"
        raw: true
    output_buffer:
      name: "raw_output"
      path: "outputs/task_pipeline_output/prioritized_gsm8k.jsonl"
      storage_type: "file"
    format:
      prompt_key: "question"
      response_key: "answer"
    # data active iterator related
    dj_process_desc: "Please compute difficulty scores for these math questions."
    agent_model_name: "qwen-max"
    clean_strategy: "iterative"
    priority_weights: 
      difficulty: -1.0  # easy-to-hard
\end{lstlisting}

\subsubsection{Dynamic Experience and Reward Shaping}
While task curation primes the model \emph{before} exploration, experience shaping refines the learning signal \emph{after} each agent-environment interaction. This is critical for RFT algorithms that rely on rich feedback, as standard rewards (e.g., binary pass/fail) are often too sparse to guide learning effectively. We demonstrate how to augment rewards with metrics for quality and diversity, transforming a sparse signal into a dense, multi-faceted one that provides clearer guidance to the trainer.

\begin{figure}
  \centering
  \includegraphics[width=.5\textwidth]{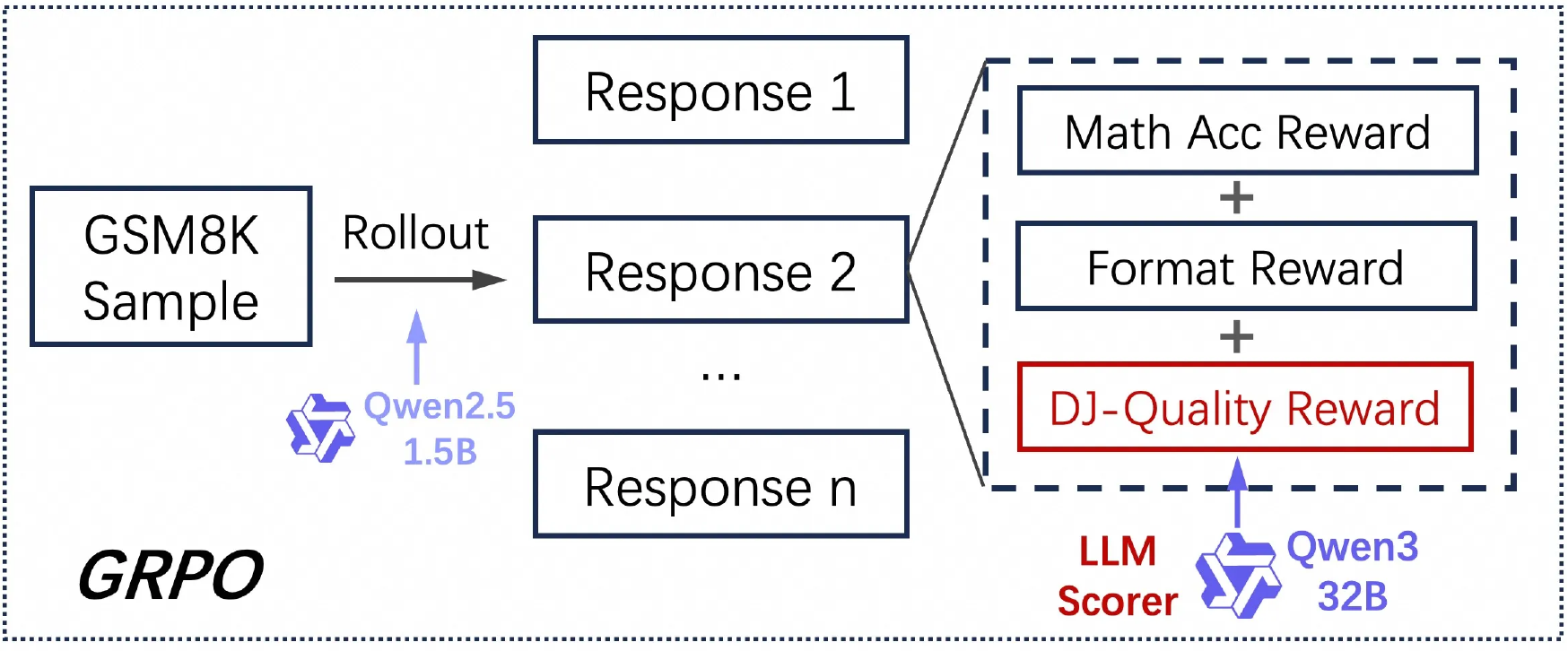}
  \caption{The enhanced math workflow with quality-reward shaping from data processor, where DJ indicates DataJuicer \cite{datajuicer}, from which more operators can be utilized to extend this worklow.}
  \label{fig:data-flow-quality-reward}
\end{figure}

\paragraph{Use Case 1: Quality Reward Augmentation.}
To encourage the model to generate high-quality responses, we can augment the base reward with a quality score. As illustrated in Figure~\ref{fig:data-flow-quality-reward}, during each RFT step, we use the data processor to evaluate the quality of each generated rollout.
For our experiment, we trained a Qwen2.5-1.5B model and used a more powerful Qwen3-32B as the scorer LLM. Specifically, we invoked the \texttt{llm\_quality\_filter} from Data-Juicer, which normalized the quality scores to the range [-0.5, 0.5] and added them to the original reward.

\begin{figure}
  \centering
  \includegraphics[width=.8\textwidth]{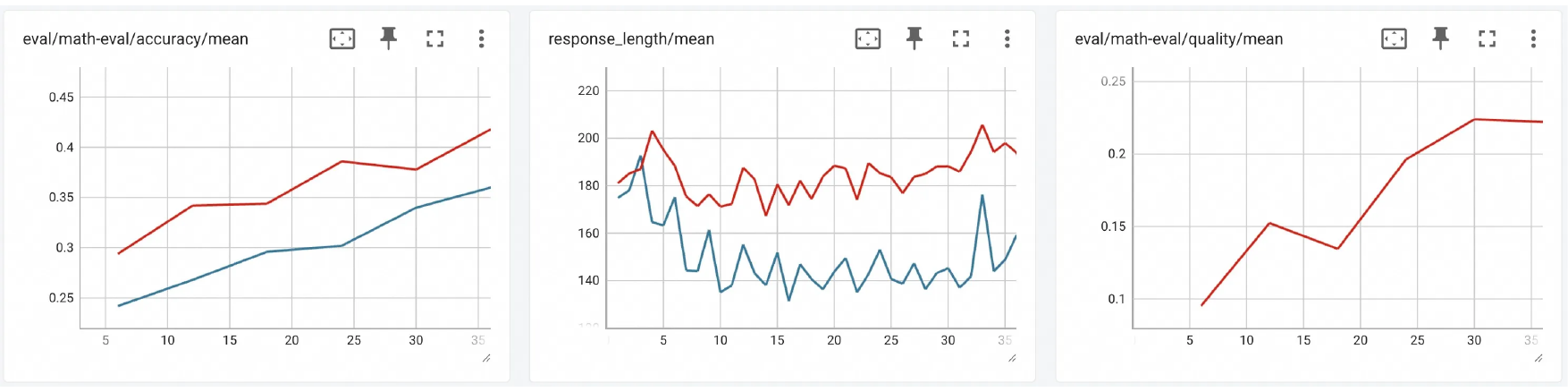}
  \caption{Experimental results for quality-reward shaping. Augmenting the reward with a quality score (red line) improves final accuracy and provides a learnable reward signal.}
  \label{fig:data-flow-quality-reward-res}
\end{figure}

Crucially, this processing is applied to the \textbf{experience buffer} at each RFT step. This allows the reward signal to adapt dynamically to the policy model's evolving capabilities, a more responsive approach than one-time static processing. With a \texttt{sync\_interval} of 3 over 36 steps on the Math-500 validation set, the results in Figure~\ref{fig:data-flow-quality-reward-res} show that: \textbf{(1)} The model with quality reward augmentation (red line) achieves higher accuracy. \textbf{(2)} The introduced quality reward itself improves over time, confirming it is a \emph{learnable} signal. \textbf{(3)} We observe a slight increase in response length, which likely reflects an inductive bias from the larger scorer model being implicitly distilled into the smaller policy model.

\begin{figure}
  \centering
  \includegraphics[width=.6\textwidth]{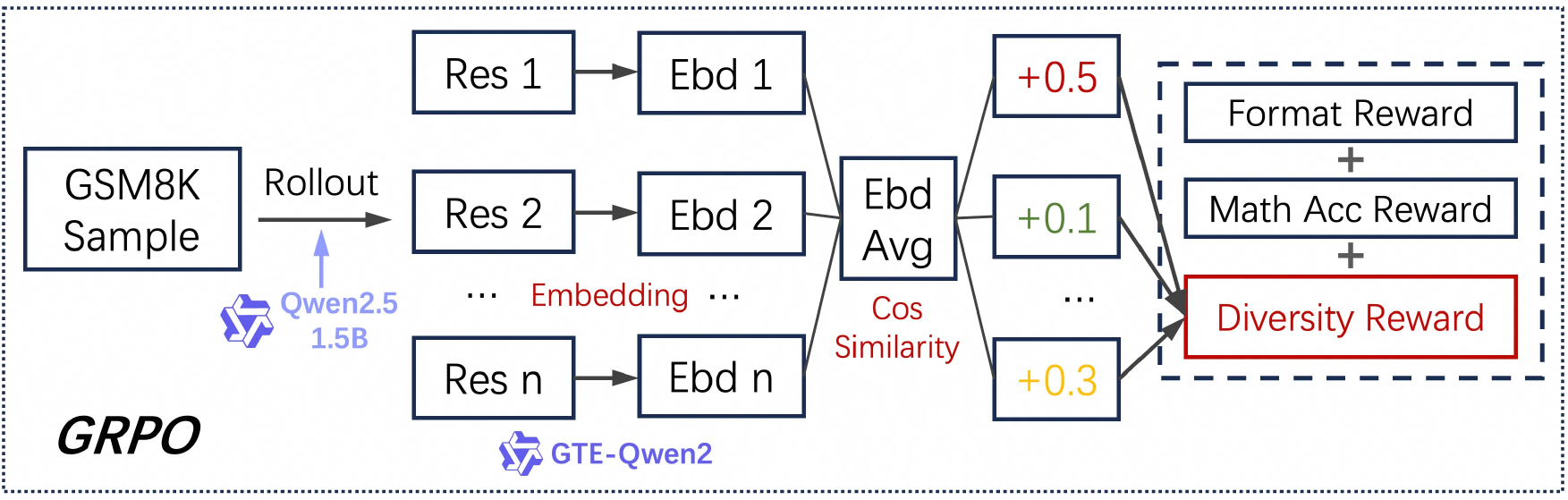}
  \caption{The enhanced math workflow with diversity-reward shaping from data processor}
  \label{fig:data-flow-diversity-reward}
\end{figure}

\paragraph{Use Case 2: Diversity Reward Augmentation.}
A common failure mode in RFT is \emph{policy collapse}, where the agent repeatedly generates similar, suboptimal responses. To counteract this, we introduce a diversity reward that encourages the explorer to explore different solution paths. As shown in Figure~\ref{fig:data-flow-diversity-reward}, we used the GTE-Qwen2-1.5B model to convert rollouts into semantic embeddings. The diversity reward was calculated based on the cosine similarity of a rollout's embedding to the mean embedding of its group, with lower similarity (i.e., higher diversity) yielding a higher reward.

\begin{figure}
  \centering
  \includegraphics[width=.8\textwidth]{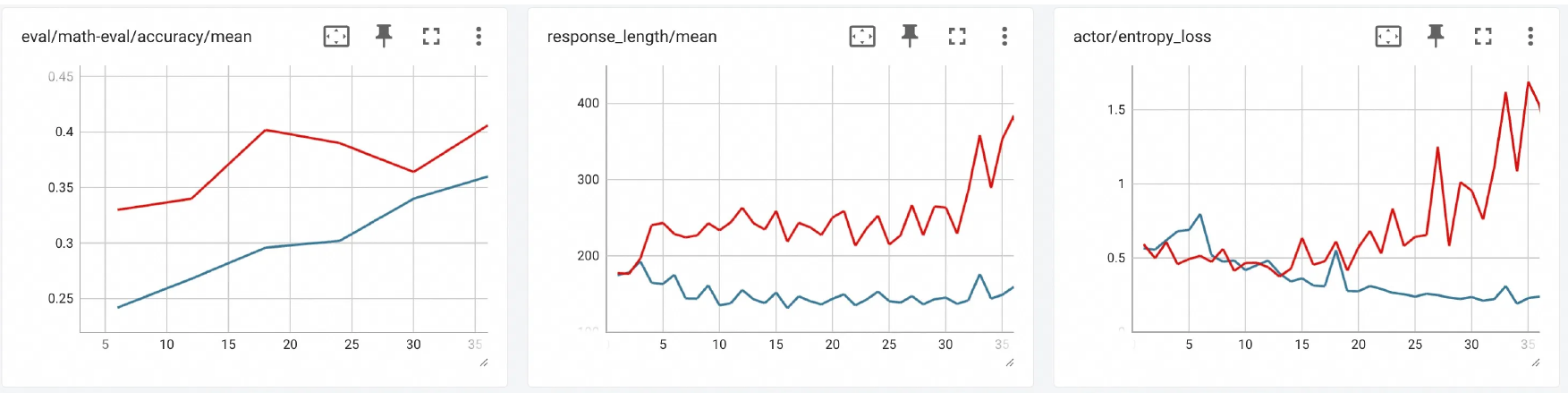}
  \caption{Experimental results for diversity-reward shaping. Rewarding diverse responses (red line) significantly improves task accuracy and maintains higher entropy.}
  \label{fig:data-flow-diversity-reward-res}
\end{figure}

To prevent exploration from becoming chaotic, we applied a simple decay schedule to the diversity reward's weight, starting at 0.5 and decaying to 0.3 over the training steps. The experiment, using the same setting as before, yielded compelling results (Figure~\ref{fig:data-flow-diversity-reward-res}): \textbf{(1)} The diversity-augmented model (red line) shows a significant performance improvement over the baseline. \textbf{(2)} The response length is consistently longer, indicating the reward encourages more elaborate answers. \textbf{(3)} Most importantly, the \textbf{actor entropy loss} remains consistently higher, providing strong evidence that the model is maintaining a healthier, more diverse exploration strategy, which helps it escape local optima.

\subsection{RFT with Human in the Loop}
\label{sec:exapme:human-loop}

\begin{figure}[h]
    \centering
    \includegraphics[width=0.85\textwidth,trim=0 620 320 0,clip]{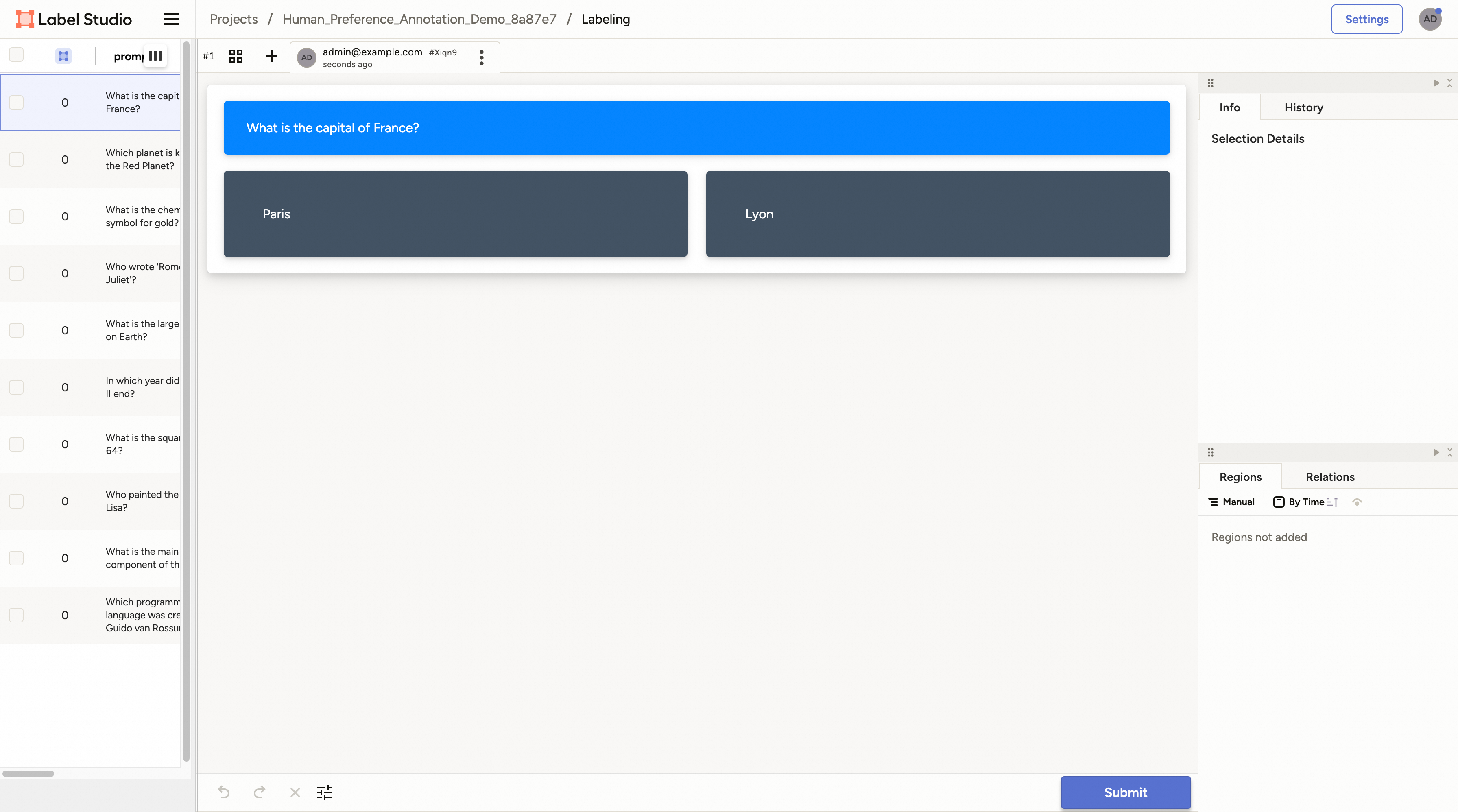}
    \caption{An interactive interface for human preference annotation.
    }
    \label{fig:human-interface}
\end{figure}

This example demonstrates the {human-in-the-loop} capability in \trinity for preference modeling. As illustrated in Listing~\ref{lst:human_annotation_config} and Figure~\ref{fig:human-interface}, our framework integrates Label Studio's annotation interface with asynchronous data pipelines through four coordinated stages:
(1) task generation: auto-creating annotation batches from model rollouts; 
(2) interactive labeling: providing UI for side-by-side response comparison; 
(3) quality control: enforcing inter-annotator agreement thresholds; and 
(4) versioned storage: tracking preference lineage in pre-defined fields like those in \texttt{DPODataModel}. 

This pipeline reflects \trinity's {bi-directional collaboration} feature (Section~\ref{sec:data-pipelines-human}), backed by timeout-aware task polling and support of atomic batch commit. 
It enables hybrid procedures where initial AI pre-screening can reduce human workload in production deployments. Annotation activities can scale across distributed teams through event-driven task routing.
The system's flexibility benefits rapid adaptation to diverse annotation protocols, allowing developers to implement custom labeling interfaces through XML-based templates or integrate third-party annotation services via unified SDK endpoints.
This capability underpins advanced use cases such as safety red-teaming datasets and online instruction tuning scenarios where human judgment remains irreplaceable for quality-critical decisions, particularly in human-centric sociocultural contexts where data quality, difficulty, and reward signals are difficult to verify logically.

\begin{lstlisting}[language=Python, label={lst:human_annotation_config}, caption=Configuration for human preference annotation.]
# Human annotation configuration
class HumanAnnotationConfig:
    """Preference annotation pipeline configuration"""
    
    def __init__(self):
        self.process = [
            {
                "human_preference_annotation_mapper": {
                    "wait_for_annotations": True,  # Block until annotations complete
                    "timeout": 3600,               # Maximum wait time in seconds
                    "prompt_key": "prompt",        # Source field for prompts
                    "answer1_key": "answer1",      # First candidate response
                    "chosen_key": "chosen"         # Selected response key
                }
            }
        ]
        
    def get_pipeline(self) -> List[Dict]:
        """Get annotation processing pipeline"""
        return self.process
\end{lstlisting}

\subsection{Low-Code Usage and Development with \trinitystudio}
\label{subsec:trinity-studio}

\begin{figure}
    \centering
    \begin{subfigure}{.6\textwidth}
        \centering
        \includegraphics[width=\textwidth]{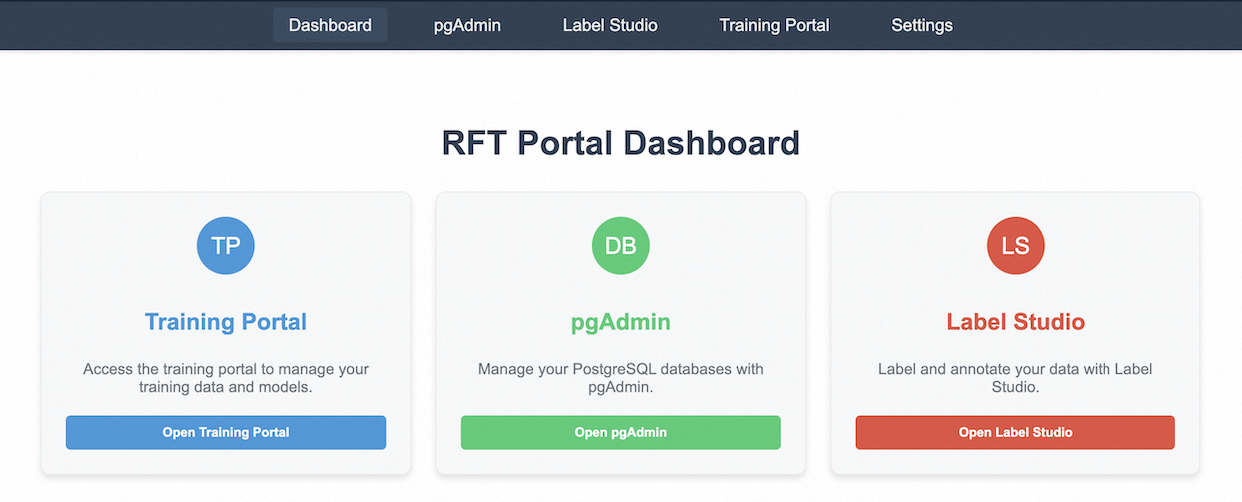}
        \caption{\trinitystudio dashboard.}
        \label{fig:trinity-studio:dashboard}
        \vspace{0.5em}
    \end{subfigure}
    \begin{subfigure}{.8\textwidth}
        \centering
        \includegraphics[width=\textwidth]{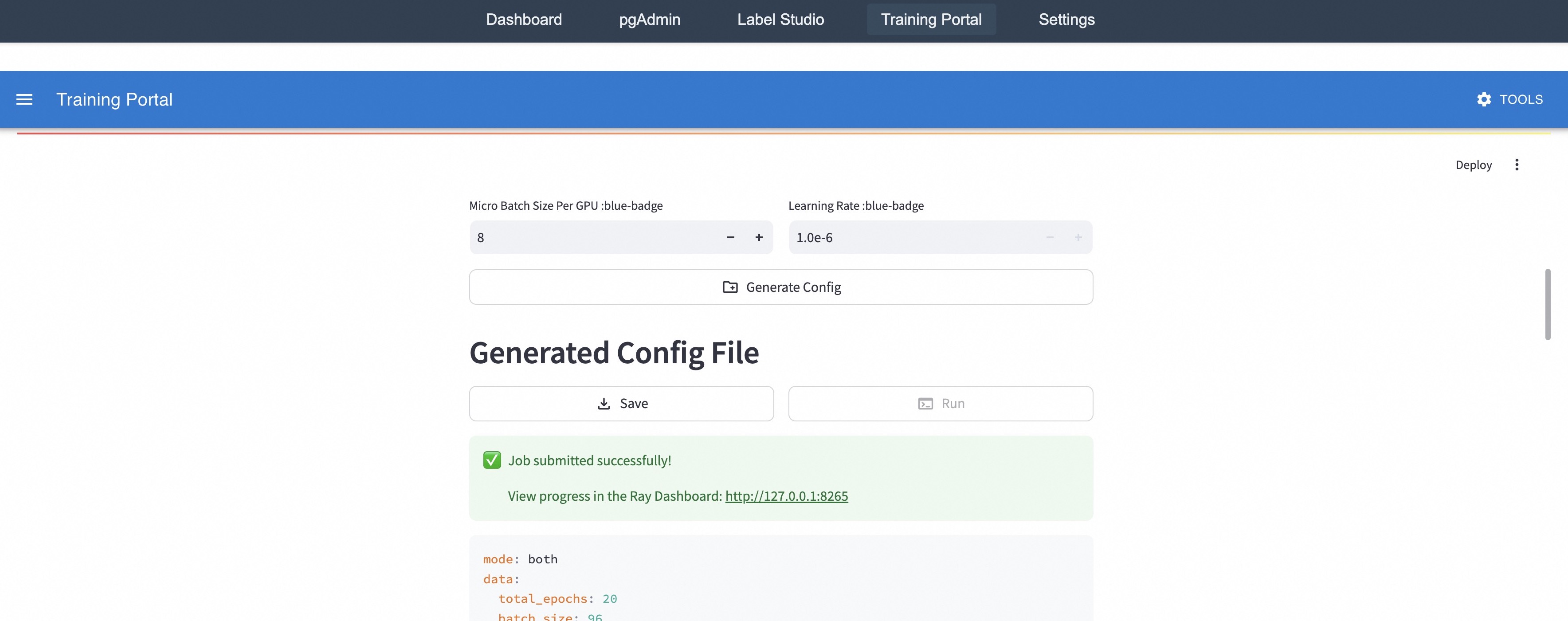}
        \caption{Start training on the ``Training Portal'' page.}
        \label{fig:trinity-studio:training}
        \vspace{0.5em}
    \end{subfigure}
    \begin{subfigure}{.8\textwidth}
        \centering
        \includegraphics[width=\textwidth]{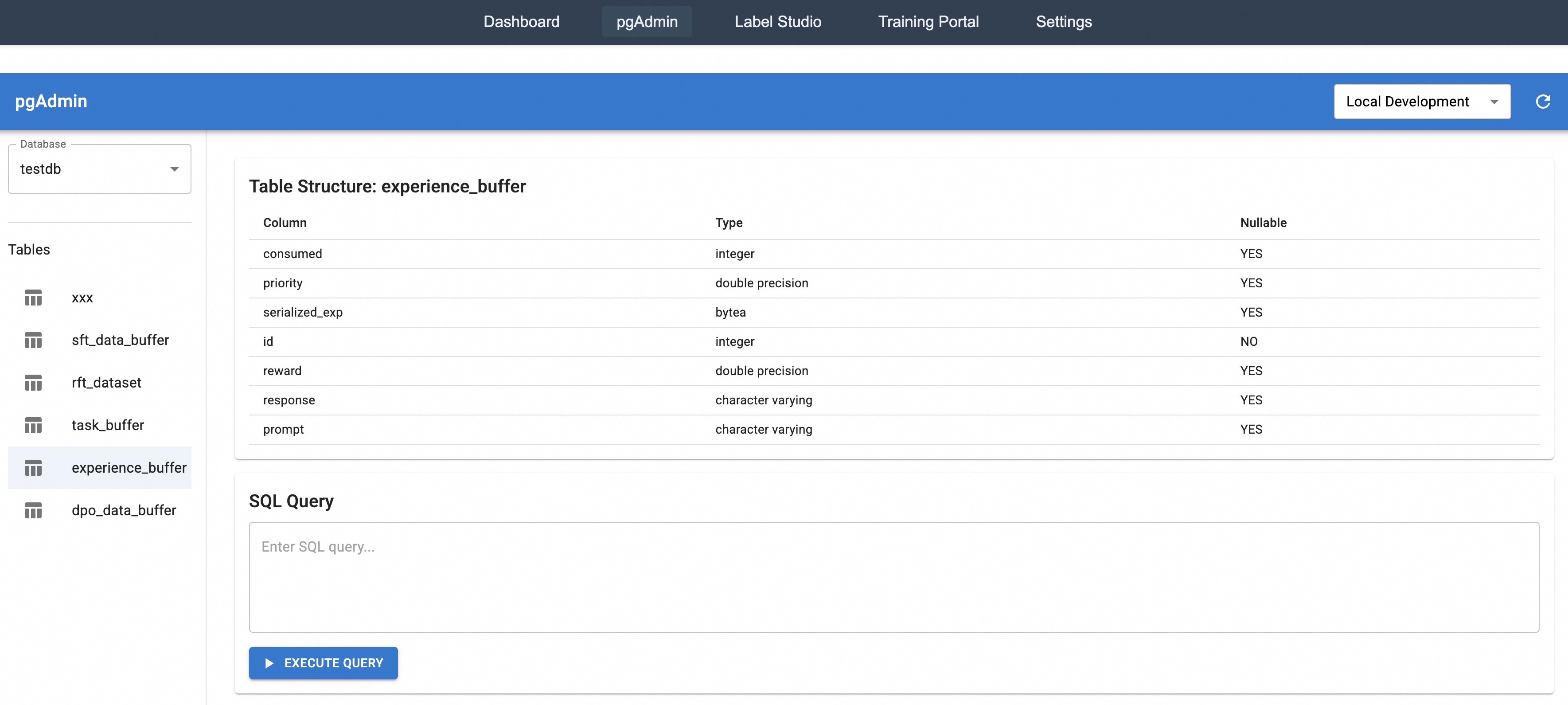}
        \caption{Manage data on the ``pgAdmin'' page.}
        \label{fig:trinity-studio:pgadmin}
        \vspace{0.5em}
    \end{subfigure}
    \begin{subfigure}{.8\textwidth}
        \centering
        \includegraphics[width=\textwidth]{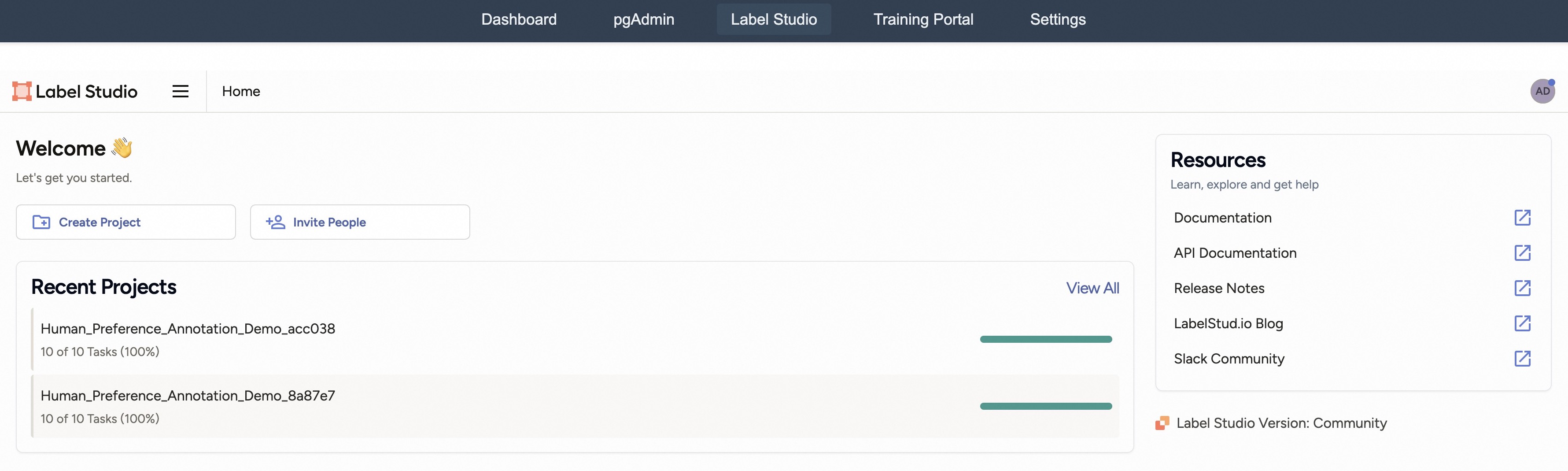}
        \caption{Process data on the ``Label Studio'' page.}
        \label{fig:trinity-studio:label-studio}
    \end{subfigure}

    \caption{Snapshots of \trinitystudio.}
    \label{fig:trinity-studio}
\end{figure}

\trinitystudio provides visual interaction for the core capabilities of \trinity, designed to bridge the gap between system complexity and user accessibility. 
As shown in Figure~\ref{fig:trinity-studio:dashboard}, its three integrated modules --- {``Training Portal''}, {``pgAdmin''}, and {``Label Studio''} --- form a cohesive interface that supports low-code usage and development with \trinity, and make it easy to monitor and track the full RFT pipeline with transparency.
\begin{itemize}
\item {``Training Portal''} (Figure~\ref{fig:trinity-studio:training}) implements configuration-to-execution procedures through declarative YAML editing with auto completion and live validation that prevents misconfigurations. 
Furthermore, the integration of runtime metrics with tools like Wandb/TensorBoard directly helps the active data optimization feature by surfacing signals such as difficulty distribution drifts and diversity metrics mentioned in Section~\ref{subsec:examples_data_processor}. This transparency ensures that users can monitor how data curation strategies impact RFT performance in real time.
\item {``pgAdmin''} (Figure~\ref{fig:trinity-studio:pgadmin}) reflects \trinity's \emph{end-to-end data transformation} capabilities by providing a visual panel for PostgreSQL-based storage.
This design benefits the \emph{versioned data lineage} requirements of RFT, particularly for scenarios involving asynchronous training (Section~\ref{sec:data-pipelines-shape}). With intuitive SQL query builders, users can easily adjust schema, audit training experiences and human annotation batches with fine-grained precision. This capability is valuable for rapid validation of active learning policies by cross-referencing training outcomes with metadata (e.g., difficulty scores and staleness in asynchronous mode).
\item {``Label Studio''} page (Figure~\ref{fig:trinity-studio:label-studio}) operationalizes \trinity's bi-directional human-AI collaboration capability (Section~\ref{sec:data-pipelines-human}).  
Utilizing the provided task polling and atomic batch commit mechanisms, users can annotate the data or experiences directly, allowing an asynchronous way to involve human feedback and to dynamically influence data curation.
\end{itemize}
By unifying these capabilities in a single UI, \trinitystudio reduces the cognitive load of managing complex RFT procedures. For example, a researcher tuning a math reasoning task could use the Training Portal to adjust difficulty scoring parameters, view the resulting distribution shifts in the pgAdmin module, and then validate human annotators' preferences in the Label Studio page. 
This end-to-end visibility can be useful for debugging and iterating RFT strategies,
and complements the programmatic APIs of \trinity while maintaining full compatibility with CLI procedures.

We implement \trinitystudio with the Singe-Spa framework \cite{singe-spa}.
The modular architecture enables custom view development through JavaScript plugins and flexible extensions for general-purpose usage.

\section{Conclusion and Next Steps}

We have presented \trinity, 
a general-purpose, flexible, scalable and user-friendly framework for reinforcement fine-tuning of large language models.
\trinity offers a path into ``the era of experience'' \cite{Silver2025},
by supporting applications in diverse scenarios with complex agent-environment interaction,
and serving as a unified platform for exploring advanced methodologies in each stage of the complete RFT pipeline, 
at both macroscopic and microscopic levels.

Further development of \trinity includes 
incorporating more advanced RL algorithms (especially off-policy or asynchronous ones),
making the choices of hyperparameters more adaptive and less reliant on manual tuning,
augmenting data pipelines with smarter sampling strategies and data processing operations,
and conducting more thorough experiments and evaluations with \trinity.

\section*{Acknowledgements}

\trinity is built upon many excellent open-source projects,
including but not limited to: 
verl \cite{verl} and PyTorch's FSDP \cite{fsdp} for LLM training;
vLLM \cite{vllm} for LLM inference;
Data-Juicer \cite{datajuicer} for data-related functionalities;
AgentScope \cite{agentscope} for agentic workflow;
and Ray \cite{ray} for distributed runtime.

\bibliographystyle{plain}
\bibliography{refs}

\end{document}